\documentclass[11pt]{article}

\usepackage[final]{acl}

\usepackage{times}
\usepackage{latexsym}

\usepackage[T1]{fontenc}

\usepackage[utf8]{inputenc}

\usepackage{microtype}

\usepackage{inconsolata}

\usepackage{graphicx}

\usepackage{amsmath,amssymb,amsfonts}
\usepackage{algorithmic}
\usepackage{textcomp}
\usepackage{kotex}
\usepackage{nicematrix} 
\usepackage{multirow} 
\usepackage{booktabs} 
\usepackage{amsmath} 
\usepackage{xcolor}
\usepackage{soul}
\usepackage{hhline}
\usepackage{pifont}
\usepackage{makecell}
\usepackage{bibentry}
\usepackage{hyperref}
\usepackage{adjustbox}
\patchcmd{\thebibliography}{\chapter*}{\clearpage\chapter*}{}{}
\usepackage{stfloats}
\usepackage{dsfont}

\newcommand{\hy}{\hat{y}}

\newcommand{\hp}{\hat{p}}

\renewcommand{\Sigma}{\mathfrak{S}}

\newcommand{\indic}[1]{\mathds{1}_{\{#1\}}}

\newcommand{\lmatch}[1]{{\cal L}_{\rm match}(#1)}

\def\eqref#1{equation~\ref{#1}}

\def\1{\bm{1}}

\DeclareMathAlphabet{\mathsfit}{\encodingdefault}{\sfdefault}{m}{sl}
\SetMathAlphabet{\mathsfit}{bold}{\encodingdefault}{\sfdefault}{bx}{n}

\DeclareMathOperator*{\argmin}{arg\,min}

\title{RA-RRG: Multimodal Retrieval-Augmented Radiology Report Generation with Key Phrase Extraction}

\author{
 \quad
{\bf Jonggwon Park}$^{1}$ \quad
{\bf Byungmu Yoon}$^{1}$ \quad
{\bf Soobum Kim}$^{1}$ \quad
{\bf Kyoyun Choi$^{2}$}\thanks{Corresponding author.} \\
\\
$^{1}$ DEEPNOID Inc., Seoul, South Korea \\ 
$^{2}$ Department of Artificial Intelligence and Data Science, Sejong University, Seoul, South Korea \\
{\tt\small jgpark@deepnoid.com, kychoi@sejong.ac.kr}
}

\begin{document}
\maketitle

\begin{abstract}

Automated radiology report generation (RRG) holds potential to reduce the workload of radiologists, 
and recent advances in multimodal large language models (MLLMs) have enabled multimodal chest X-ray (CXR) report generation.
However, existing MLLMs are 
computationally expensive, require large-scale training data, and may produce hallucinated content,
limiting their practical deployment.
To address these limitations, we propose RA-RRG, a retrieval-augmented RRG framework that
combines multimodal retrieval with 
large language models (LLMs)
to generate radiology reports while reducing hallucinations and computational demands.
RA-RRG uses LLMs to extract clinically essential key phrases from radiology reports and retrieves relevant phrases given an input image.
By conditioning LLMs on the retrieved phrases, RA-RRG effectively suppresses hallucinations while maintaining strong report generation performance.
Experiments on the MIMIC-CXR and IU X-ray datasets show state-of-the-art results on CheXbert metrics and competitive RadGraph F1 scores compared to MLLMs.
Furthermore, RA-RRG naturally generalizes to multi-view RRG 
by aggregating phrases retrieved from multiple images, 
highlighting its broad applicability to real-world clinical scenarios.
Code is available at \href{https://github.com/deepnoid-ai/RA-RRG}{https://github.com/deepnoid-ai/RA-RRG}.

\end{abstract}

\section{Introduction}
\label{sec:intro}

Automated radiology report generation (RRG) has the potential to substantially reduce the workload of radiologists by translating medical images into 
textual descriptions.
Recent advances in large language models (LLMs) have further expanded this potential, particularly through multimodal models that jointly process images and text, including chest X-rays (CXR) \cite{llava-rad, maira1, medpalm-m, medgemini2d}.
Despite their strong performance, such multimodal LLMs (MLLMs) typically require extensive computational resources and large-scale fine-tuning, 
which hinders their adoption in clinical settings.

Retrieval-augmented generation (RAG) \cite{rag} offers a promising alternative by enhancing generation through external knowledge retrieval.
In CXR RRG, prior work has explored multimodal retrieval-based approaches \cite{cxr-repair, cxr-redone}, which retrieve similar reports or sentences based on an input image.
However, radiology reports often describe multiple co-occurring findings, and naively retrieved text may include information that is irrelevant or even contradictory to the given image.
This issue is exacerbated when sentences from different reports are combined \cite{transq, teaser}.
Moreover, radiology reports frequently contain comparative statements referring to prior examinations.
When only a single image is available, we define such comparative content as \textit{comparative hallucinations}, as it is unsupported by the input.

\begin{figure*}[ht]
    \begin{center}
    \scalebox{0.95}{
        \includegraphics[width=\linewidth]{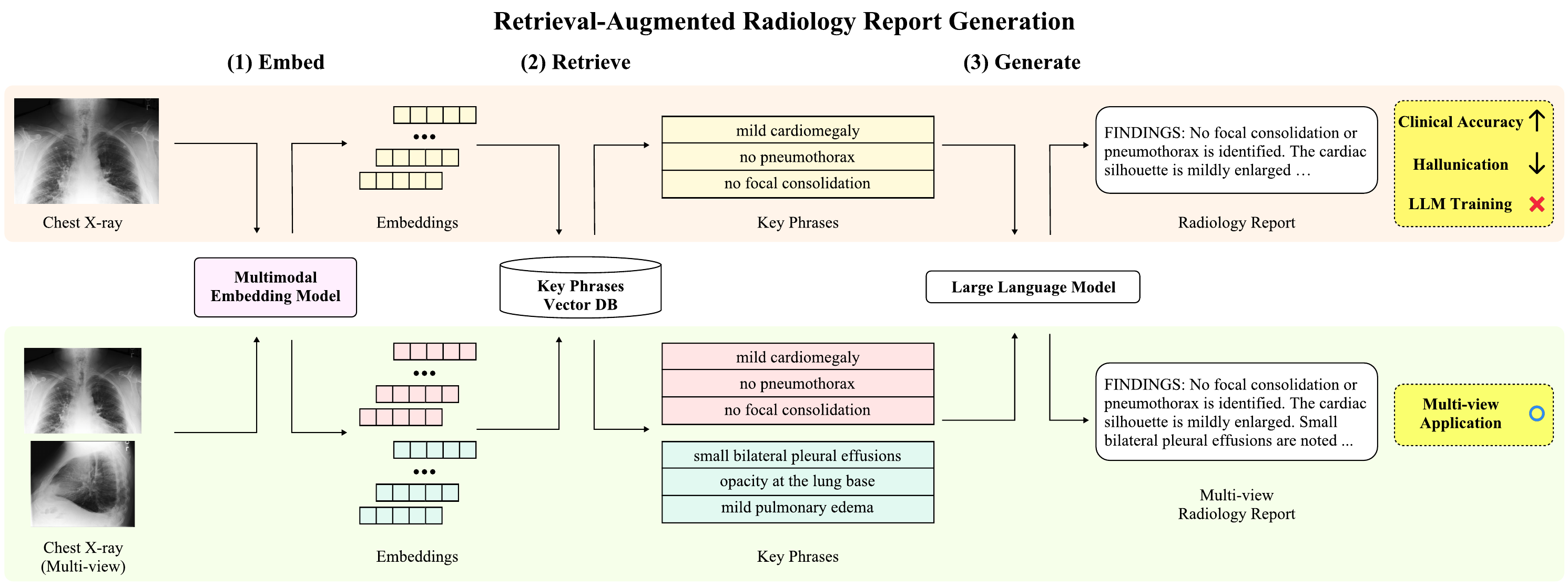}
    }
    \vspace{-0.5cm}
    \end{center}
    \caption{
    Overview of RA-RRG framework.
    Given a chest X-ray, a multimodal retriever selects clinically essential key phrases, which are then provided to an LLM to generate hallucination-suppressed reports without LLM training.
    The same pipeline naturally extends to multi-view inputs.
    }
    \vspace{-0.3cm}
    \label{fig:intro}
\end{figure*}

To address these challenges, we propose \textbf{RA-RRG}, a \textbf{R}etrieval-\textbf{A}ugmented \textbf{RRG} framework that combines
LLMs with multimodal retrieval without requiring any LLM fine-tuning.
Building on RadGraph \cite{radgraph}, we use an LLM to extract clinically essential key phrases from reports while explicitly excluding undesired content such as comparisons with prior studies, 
resulting in comparative hallucination-free key phrases aligned with the visual evidence.
Given an input image, RA-RRG retrieves relevant key phrases and conditions an LLM on these 
image-consistent phrases to generate accurate and reliable reports without any LLM training.
Experimental results on MIMIC-CXR and IU X-ray demonstrate that RA-RRG achieves strong performance on CheXbert metrics and competitive RadGraph F1 scores, while requiring only 18 GPU-hours for training, compared to over 200 GPU-hours for comparable MLLMs.
Furthermore, the proposed framework naturally extends to multi-view RRG by aggregating 
phrases retrieved independently from multiple images.
Figure \ref{fig:intro} provides an overview of the RA-RRG framework.

Our main contributions are summarized as follows:
(1) We propose RA-RRG, a retrieval-augmented RRG framework that produces clinically accurate radiology reports without LLM fine-tuning.
(2) RA-RRG effectively suppresses both comparative and object-level hallucinations.
(3) We demonstrate that RA-RRG generalizes well to multi-view settings, highlighting its broad applicability in real-world scenarios.

\section{Related Works}

\subsection{Retrieval Augmented Generation}
While LLMs have achieved human-level knowledge in various fields, they still suffer from outdated knowledge and hallucinations \cite{hallucination}. 
Combining retrieval-augmented generation (RAG) with LLMs \cite{rag, rag-survey} addresses these issues by retrieving information from an external database based on the query, allowing for updates without retraining the LLM.

Recent advances in MLLMs have expanded RAG to multimodal applications, including text-to-image generation \cite{reimagen, ramlm}, image captioning \cite{sarto2022retrieval, smallcap, evcap}, and video captioning \cite{ravideocap}. 
In this study, we apply a multimodal RAG approach to generate radiology reports 
by retrieving text data with embeddings aligned to 
CXR images.

\subsection{Radiology Report Generation}
Automated 
RRG
has been actively studied in recent years.
With the advent of 
MLLMs, 
CXR-focused systems such as LLaVA-Rad \cite{llava-rad}, CheXagent \cite{chexagent}, MAIRA-1 \cite{maira1}, MAIRA-2\cite{maira2}, and M4CXR \cite{m4cxr} 
have demonstrated 
report generation capabilities.
General-purpose medical foundation models, including Med-Gemini \cite{medgemini2d} and MedPaLM-M \cite{medpalm-m}, also support CXR report generation but require substantial computational resources and large-scale training 
data.

Retrieval-based approaches mitigate these limitations by leveraging existing reports.
TranSQ \cite{transq} framed RRG as a set prediction problem, while Teaser \cite{teaser} introduced topic-wise retrieval with contrastive learning.
CXR-RePaiR \cite{cxr-repair} and CXR-ReDonE \cite{cxr-redone} aligned CXR images and reports using CLIP- and ALBEF-based objectives, respectively, and CXR-RAG \cite{cxr-rag} combined retrieval with a pre-trained LLM for report generation.
\citet{bootstrappingLLM} further improved retrieval-based RRG through in-domain adaptation and contrastive ranking with structured decoding.

Another line of work leverages RadGraph \cite{radgraph} to represent reports as structured clinical knowledge.
Style-aware RRG \cite{style-aware-rrg} serialized RadGraph outputs to model radiologist-specific styles, while FactMM-RAG \cite{fact-aware-mrag} focused on pathology-centric factual extraction with contrastive learning.
In contrast, our approach integrates an LLM to refine RadGraph outputs into hallucination-suppressed key phrases for retrieval-augmented RRG.

\begin{figure*}[!t]
    \begin{center}
    \scalebox{0.95}{
        \includegraphics[width=\linewidth]{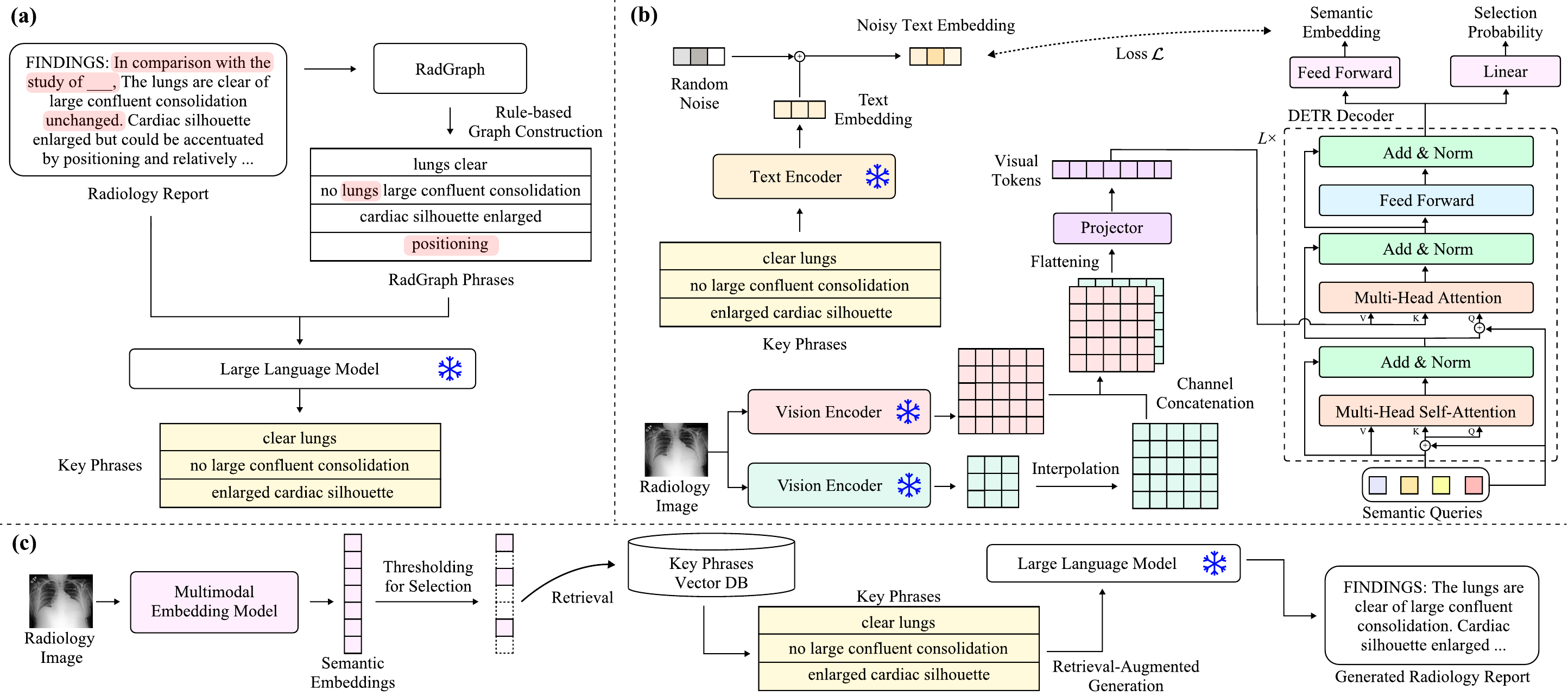}
    }
    \vspace{-0.5cm}
    \end{center}
    \caption{
    (a) Key phrase extraction using an LLM. 
    (b) The multimodal retriever architecture. 
    (c) Inference process of RA-RRG.
    }
    \vspace{-0.3cm}
    \label{fig:method}
\end{figure*}

\section{Methods}
\label{sec:methods}

\subsection{LLM-Based Key Phrase Extraction}\label{subsec:key_phrase_extraction}

Most prior retrieval-based RRG approaches treat the entire radiology report as the retrieval target \cite{cxr-repair, cxr-rag} or split reports into sentence-level segments \cite{transq, teaser}.
However, both strategies suffer from co-occurrence issues, as multiple independent findings may coexist within a single text, and reports often include extraneous information such as doctor names or user metadata.
To better utilize radiology reports for training, we decompose reports into minimal clinically meaningful phrases while removing unnecessary content.
Specifically, we apply RadGraph \cite{radgraph} to the \textit{FINDINGS} section to extract entities and relations, and construct RadGraph phrases that 
represent the key clinical findings in the report
(see Appendix~\ref{appendix:key_prhase}).

Despite its effectiveness, RadGraph may produce fragmented graphs and does not explicitly address hallucination-prone expressions such as comparative statements involving prior studies (e.g., \textit{improved} or \textit{unchanged}), which are unsupported in single-image RRG.
To address this limitation, we employ an LLM for key phrase extraction, inspired by prior work on LLM-based knowledge graph extraction \cite{hipporag}, and refine RadGraph outputs into key phrases while excluding hallucination-indicative terms.
Because general-purpose LLMs may omit domain-specific clinical details when processing raw text alone, we provide both the original report and RadGraph-derived structures as input, 
with examples and prompt templates shown in Figure~\ref{fig:method} (a) and Appendix Figure~\ref{fig:key-phrase-extraction-prompt}.

\subsection{Multimodal Retriever}

\subsubsection{Model Architecture}

To train a multimodal retrieval model using 
images and their corresponding lists of key phrases,
our model builds on the architecture of TranSQ \cite{transq}.
TranSQ adapts the training approach of DETR \cite{detr} 
for sentence-level retrieval. 
Our model consists of a vision encoder, a DETR decoder, and a text encoder, as illustrated in Figure \ref{fig:method} (b).

\paragraph{Vision encoder}
We leverage complementary strengths of pre-trained vision encoders obtained from different pre-training paradigms, including vision–language pre-training (e.g., CLIP \cite{clip}) and unimodal self-supervised learning (e.g., DINOv2 \cite{dinov2}).
Rather than relying on a single encoder, we fuse visual features from multiple encoders using channel concatenation \cite{eagle}.
All encoders follow a Vision Transformer (ViT) \cite{vit} architecture and output sequences of visual tokens.
To address differences in sequence length, we reshape the token sequences into 2D feature maps, apply spatial interpolation to align their resolutions, and concatenate them along the channel dimension to form a unified visual representation.

\paragraph{Text encoder}
During training, 
the text encoder converts key phrases, 
extracted from the report corresponding to each training image, 
into text embeddings.
Since the text encoder is kept frozen as in \cite{transq}, 
the resulting text embeddings for each training image 
remain fixed, which can lead to overfitting.
Inspired by NEFTune \cite{neftune}, which adds random noise to 
embeddings during LLM fine-tuning, 
we similarly inject noise into text embeddings during training.
The noise $\epsilon$ is sampled from a uniform distribution in the range [-1, 1] and scaled by
$1/\sqrt{d}$,
where $d$ is the embedding dimension.
We also apply 
L2 normalization to the text embeddings.
For inference, we construct a vector database of embeddings from all key phrases in the training set to facilitate retrieval.

\paragraph{DETR decoder}
Similar to TranSQ, we adopt the original DETR decoder structure. 
The visual token sequence from the vision encoder 
serves as the encoder input, while $N$ query embeddings are decoded in parallel through self-attention and encoder-decoder attention. 
To match the feature dimension of the visual token sequence with 
that of the decoder,
we apply a linear projection layer.
A selection classifier, 
implemented as a linear layer, produces
selection logits, and semantic embeddings are 
generated by a three-layer
feed-forward network with ReLU activation. 
Each semantic embedding is then L2-normalized.

\subsubsection{Loss Function}

\paragraph{Phrase matching loss}
We optimize the retriever using a phrase matching loss.
Similar to DETR,
TranSQ applies the Hungarian algorithm \cite{hungarian} based on selection probabilities and the similarity between semantic 
and text embeddings.
Let $y$ represent the ground truth set of key phrases.
$\hat{y}=\{\hat{y}_i\}^N_{i=1}$ consists of $N$ predictions. 
We set $N$ to exceed the number of key phrases 
and pad $y$ with empty elements ($\varnothing$) to form equal-sized sets.
The Hungarian algorithm then finds an optimal permutation
$\sigma \in \Sigma_N$ that minimizes the matching cost:
\vspace{-0.3cm}
\begin{equation}\label{eq:hungarian}
\hat{\sigma} = \argmin_{\sigma\in\Sigma_N} \sum_{i}^{N} \lmatch{y_i, \hy_{\sigma(i)}}
\end{equation}

Matching cost $\lmatch{y_i, \hy_{\sigma(i)}}$ is 
computed as the negative sum
of selection probability $\hat{p}_{\sigma(i)}$
(scaled by
$\mu$)
and cosine similarity ${\cal L}_{\text{sim}}$ 
between text embedding $v_i$ and semantic embedding $\hat{v}_{\sigma(i)}$:
\vspace{-0.3cm}
\begin{multline}
\lmatch{y_i, \hy_{\sigma(i)}}= \\ 
-\mu \indic{y_i\neq \varnothing}\hp_{\sigma(i)}
-\indic{y_i\neq \varnothing} {\cal L}_{sim}(v_i, \hat{v}_{\sigma(i)})
\end{multline}

Given the optimal assignment,
we compute the selection loss $ {\cal L}_{cls}$ using 
distribution-balanced loss \cite{dbloss} 
with binary classification labels $c_i = \indic{y_i\neq \varnothing}$,
and a negative cosine similarity loss to align matched text and semantic embeddings.
The overall phrase matching loss is:
\vspace{-0.3cm}
\begin{align}
\mathcal{L}_{\rm PM}(y, \hat{y})
&= \sum_{i=1}^{N} \Big[
\mathcal{L}_{cls}(c_i, \hat{p}_{\hat{\sigma}(i)}) \notag\\
&\hspace{-2.0em} + \indic{y_i \neq \varnothing\}}
\big(1 - \mathcal{L}_{sim}(v_i, \hat{v}_{\hat{\sigma}(i)})\big)
\Big]
\end{align}

\paragraph{In-batch semantic contrastive loss}
${\cal L}_{sim}$
aligns matched semantic and text embeddings
but does not discourage non-matching embeddings from becoming similar.
Thus,
we adopt a CLIP-style symmetric contrastive loss \cite{clip},
which pulls matched embedding pairs closer while pushing mismatched pairs apart.

Given a mini-batch of size $B$,
we construct a set of positive pairs
$E = \{(v_i^b, \hat{v}^b_{\hat{\sigma}(i)}) \,|\, y_i^b \neq \varnothing, \, b = 1, \dots, B\}$
from the Hungarian matching,
where $v_i^b$ is the $i$-th text embedding and $\hat{v}^b_{\hat{\sigma}(i)}$ is its matched semantic embedding in batch $b$.
We flatten all pairs across the batch and index them as $k = 1, \dots, |E|$.
Let $v_k$ and $\hat{v}_k$ denote the text and semantic embeddings of the $k$-th pair, respectively.
The cross-modal similarity matrix is defined as:
\vspace{-0.2cm}
\begin{equation}
Z_{kl} = v_k^\top \hat{v}_l
\end{equation}

Rather than using hard one-hot targets that treat all non-matched pairs as negatives,
we construct soft contrastive targets from intra-modal self-similarities
$S^v_{kl} = v_k^\top v_l$ and $S^{\hat{v}}_{kl} = \hat{v}_k^\top \hat{v}_l$:
\vspace{-0.2cm}
\begin{equation}
q_{kl} = \mathrm{softmax}_l \left( (S^v_{kl} + S^{\hat{v}}_{kl}) / 2 \right)
\end{equation}
This ensures that semantically similar pairs in both modalities are not heavily penalized.
The loss is defined as a symmetric cross-entropy:
\vspace{-0.2cm}
\begin{multline}
{\cal L}_{\rm SC} = \frac{1}{2|E|} \sum_{k=1}^{|E|} \Big[ H\big(q_{k,:},\, \mathrm{softmax}(Z_{k,:})\big) \\
+ \, H\big(q_{:,k},\, \mathrm{softmax}(Z_{:,k})\big) \Big]
\end{multline}
where $H(p, r) = -\sum_l p_l \log r_l$ denotes the cross-entropy.
The total training objective is:
\vspace{-0.2cm}
\begin{equation}
{\cal L} = \sum_{b=1}^B {\cal L}_{\rm PM}(y^b, \hat{y}^b) + \lambda {\cal L}_{\rm SC}(E).
\end{equation}

\subsection{Multimodal Retrieval-Augmented RRG}
\subsubsection{Key Phrase Retrieval}
To generate a radiology report from an image, we compute $N$ selection probabilities and the corresponding semantic embeddings. 
Only embeddings with probabilities above a set threshold 
are used for key phrase retrieval.
The retrieval target is
a vector database of text embeddings, built from 
the full set of 
key phrases gathered from the training dataset. 
Matching 
each semantic embedding to its nearest text embedding yields
a list of key phrases that describe the image.

\subsubsection{Radiology Report Generation with LLM}\label{subsec:mrg_llm}
The final step of RA-RRG uses an LLM to generate a complete radiology report from retrieved key phrases.
Since these phrases are not full sentences, the LLM integrates their content to produce a coherent and comprehensive report.
To ensure desired report qualities, such as hallucination suppression, we provide task-specific instructions together with the key phrases as input to the LLM; the prompt is shown in Figure~\ref{fig:RAG-prompt} in the Appendix.

Using an LLM  
broadens the applicability of RA-RRG beyond single-image report generation.
The framework 
extends to multi-view and follow-up scenarios by extracting key phrases from each image independently and providing them, 
along with contextual information such as view position, to the LLM.
Unlike MLLM-based approaches that require architectural modifications 
to handle multiple images \cite{maira2}, 
RA-RRG enables unified report generation in a 
straightforward manner.
\section{Experiments}
\label{sec:experiments}

\subsection{Datasets}\label{subsec:datasets}

\subsubsection{Training dataset}
For training and validation, we use the 
\textbf{MIMIC-CXR}
dataset \cite{mimic-cxr, mimic-cxr-jpg}, which contains 
paired chest X-ray images and radiology reports.
Using the official MIMIC-CXR codebase\footnote{\url{https://github.com/MIT-LCP/mimic-cxr}}
, we extract only the \textit{FINDINGS} section from each report and follow the official data split.
We retain studies with non-empty RadGraph phrases and key phrases, excluding cases without clinically meaningful reports.
After filtering, the dataset consists of 269,241 training images, 2,113 validation images, and 3,858 test images, using both frontal (PA/AP) and lateral views for training.
On average, each image is associated with 7.16 key phrases, and the training set contains 243,064 unique key phrases, reflecting substantial redundancy across reports.

\subsubsection{Test datasets}

Following the official split of 
\textbf{MIMIC-CXR},
we use 3,858 images
for single-view RRG evaluation. 
Unlike training and validation, we retain all 3,858 test images, including those with empty RadGraph or key phrases, to ensure fair comparison with prior work.
For external evaluation, we use the IU X-Ray dataset \cite{iu-xray}.
Following the setting of PromptMRG \cite{promptmrg}, we evaluate on a publicly available\footnote{\url{https://github.com/jhb86253817/PromptMRG}} subset of 4,168 images, where frontal and lateral images are treated as independent samples and normal cases are downsampled to a 10\% ratio.

Multi-view RRG is evaluated using both frontal and lateral images, following the test protocol of MAIRA-2.
Among the 3,858 MIMIC-CXR test images, 2,461 are frontal views; when multiple lateral images exist, one is selected randomly, and cases without a lateral view are evaluated using the frontal image only.

\subsection{Evaluation Metrics}

We evaluate both natural language generation (NLG) quality and clinical efficacy, using publicly available implementations\footnote{Links to the evaluation tools are provided in Appendix \ref{app:imp_details}}.
For NLG evaluation, we report ROUGE-L \cite{rouge} and BLEU scores (BLEU-1, BLEU-4) \cite{bleu}.
Clinical efficacy is assessed using CheXbert \cite{chexbert}, which labels reports across 14 observation classes. We binarize the labels by treating all non-positive labels as negative and report micro-F1, macro-F1, and example-based F1 scores \cite{cvt2dis.}.
We additionally report RadGraph F1 \cite{radgraph-f1}, which evaluates clinical correctness based on entities and relations extracted by RadGraph \cite{radgraph}.
To assess comparative hallucination, we follow \cite{cxr-redone} and measure the frequency of comparison-related keywords (e.g., unchanged, earlier, remain) and the proportion of reports containing them.
We additionally evaluate object-level hallucination by computing RadGraph precision, recall, and F1 separately for entities and relations.

\subsection{Implementation Details}\label{subsec:impl_details}
Based on an exploration of various vision encoder structures
(see Appendix \ref{subsec:ablation} for details),
we combine two image encoders made available by \citet{chexpertplus},
namely XrayDINOv2 and XrayCLIP.
For the text encoder we employ MPNet (`all-mpnet-base-v2') \cite{sbert} with 
embedding dimension $d=768$.
During multimodal retriever training,
both vision and text encoder parameters are frozen.
The parameters of DETR decoder are randomly initialized, with the number of 
query embeddings $N$ set to 50 and the number of decoder layers $L$ set to 6.
The model dimension of the DETR decoder and the dimension of the semantic embeddings are set to the same value of 768.
In the Hungarian algorithm, we set the selection probability ratio $\mu$ to 0.5. 
We set the in-batch contrastive loss ratio $\lambda$ to 0.1, and the selection probability threshold for semantic embedding retrieval to 0.4.
For key phrase extraction described in
Section \ref{subsec:key_phrase_extraction}, we utilize `Llama-3.1-70B-Instruct'
, abbreviated as Llama 70B \cite{llama3}.
When generating radiology reports in the final step,
we employ OpenAI’s GPT-4o \cite{gpt4o}
as the LLM.

\section{Results}

\subsection{Single-View RRG} \label{subsec:single-view}

\begin{table*}[t]
\caption{
Results of single-view RRG evaluation on the MIMIC-CXR test set (\textit{FINDINGS} section).
Compared methods include
METransformer \cite{metransformer},
PromptMRG \cite{promptmrg},
Med-PaLM M \cite{medpalm-m},
MAIRA-1 \cite{maira1},
LLaVA-Rad \cite{llava-rad},
M4CXR \cite{m4cxr},
TranSQ \cite{transq},
DCL \cite{dcl},
I3+C2FD \cite{bootstrappingLLM},
and MCA-RG \cite{mca-rg}.
$^*$ denotes results from PromptMRG; $^\dagger$ uses CheXpert labeling; $^\ddagger$ treats uncertain as positive.
Best values are in bold.
}
\centering
\small
\setlength{\tabcolsep}{3pt}
\renewcommand{\arraystretch}{1.12}
\label{tab:single-image-comparison-MIMIC}

\begin{adjustbox}{max width=\textwidth}
\begin{tabular}{c|c|ccc|c|ccc}
\Xhline{1pt}
\multirow{2}{*}{Type} & \multirow{2}{*}{Model}
& \multicolumn{3}{c|}{CheXbert}
& \multirow{2}{*}{\makecell{RadGraph \\ F1}}
& \multicolumn{3}{c}{NLG Metrics} \\
\cline{3-5} \cline{7-9}
 & & micro-F1 & Macro-F1 & example-F1 & & ROUGE-L & BLEU-1 & BLEU-4 \\
\hline\hline

\multirow{6}{*}{Generation}
 & METransformer$^\dagger$ & - & - & 31.1 & - & 29.1 & 38.6 & 12.4 \\
 & PromptMRG & - & 38.1 & 47.6 & - & 26.8 & 39.8 & 11.2 \\
 & Med-PaLM M 84B & 53.6 & 39.8 & - & \textbf{26.7} & 27.3 & 32.3 & 11.3 \\
 & MAIRA-1 & 55.7 & 38.6 & - & 24.3 & 28.9 & 39.2 & 14.2 \\
 & LLaVA-Rad & 57.3 & 39.5 & - & - & \textbf{30.6} & 38.1 & \textbf{15.4} \\
 & M4CXR & 58.1 & 38.8 & 50.2 & 21.7 & 28.4 & 33.3 & 10.2 \\
\hline

\multirow{5}{*}{Retrieval}
 & TranSQ$^\dagger$$^\ddagger$ & 51.9 & - & - & - & 28.6 & \textbf{42.3} & 11.6 \\
 & DCL$^*$ & - & 28.4 & 37.3 & - & 28.4 & - & 10.9 \\
 & I3+C2FD$^\dagger$ & - & - & 47.3 & - & 29.1 & 40.2 & 12.8 \\
 & MCA-RG & - & 33.5 & 40.8 & - & 30.0 & 41.1 & 12.8 \\
\cline{2-9}
 & RA-RRG & \textbf{58.5} & \textbf{41.7} & \textbf{50.7} & \textbf{26.7} & 24.9 & 37.9 & 8.0 \\
\Xhline{1pt}
\end{tabular}
\end{adjustbox}
\end{table*}

\begin{table*}[t]
\caption{
Results of single-view RRG evaluation on the IU X-Ray dataset.
The test setting follows PromptMRG, and evaluation results of other models
are referenced from the same source.
Best values are highlighted in bold.
}
\centering
\small
\setlength{\tabcolsep}{3pt}
\renewcommand{\arraystretch}{1.12}
\label{tab:single-image-comparison-IU}

\begin{adjustbox}{max width=\textwidth}
\begin{tabular}{c|l|ccc|c|ccc}
\Xhline{1pt}
\multirow{2}{*}{Type} & \multirow{2}{*}{Model}
& \multicolumn{3}{c|}{CheXbert}
& \multirow{2}{*}{\makecell{RadGraph \\ F1}}
& \multicolumn{3}{c}{NLG Metrics} \\
\cline{3-5} \cline{7-9}
 & & micro-F1 & Macro-F1 & example-F1
 & & ROUGE-L & BLEU-1 & BLEU-4 \\
\hline\hline

\multirow{4}{*}{Generation}
 & R2Gen \cite{r2gen}             & - & 7.1  & 13.6 & - & 25.3 & 32.5 & 5.9 \\
 & CvT2DistilGPT2 \cite{cvt2dis.} & - & 15.5 & 16.8 & - & 27.7 & 38.3 & 8.2 \\
 & RGRG  \cite{rgrg}              & - & 18.7 & 18.0 & - & 18.0 & 26.6 & 6.3 \\
 & PromptMRG \cite{promptmrg}     & - & 24.6 & 21.1 & - & \textbf{28.1} & \textbf{40.1} & \textbf{9.8} \\
\hline

\multirow{3}{*}{Retrieval}
 & M2KT \cite{m2kt} & - & 15.1 & 14.5 & - & 26.1 & 37.1 & 7.8 \\
 & DCL \cite{dcl}   & - & 17.7 & 16.2 & - & 26.7 & 35.4 & 7.4 \\
\cline{2-9}
 & RA-RRG             & 36.5 & \textbf{26.6} & \textbf{24.4}
                       & 30.8 & 27.2 & 36.3 & 6.7 \\
\Xhline{1pt}
\end{tabular}
\end{adjustbox}
\end{table*}

Table \ref{tab:single-image-comparison-MIMIC} reports the single-view RRG results on MIMIC-CXR.
We compared our method with state-of-the-art generative RRG models and retrieval-based approaches.
Our method achieved state-of-the-art performance on all CheXbert metrics, with a Macro-F1 of 41.7, outperforming Med-PaLM M 84B (39.8) by 1.9 points, while also yielding strong micro-F1 and example-F1 scores.
Notably, these results were obtained without LLM fine-tuning, surpassing even fine-tuned MLLMs.
On RadGraph F1, our model achieved 26.7, matching the previous best result.
This indicated that RA-RRG effectively captured clinically relevant entities and relations through key phrase retrieval.
We also evaluate RA-RRG with open-source LLMs (Llama-3.1-8B and 70B) for report generation; results in Appendix Table~\ref{tab:Ablation} show comparable performance, confirming that the framework is not dependent on proprietary LLMs.

Table \ref{tab:single-image-comparison-IU} shows the held-out evaluation results on the IU X-Ray dataset.
We followed the test setting of \citet{promptmrg}, referencing evaluation results of other models from the same source.
RA-RRG achieved the highest CheXbert Macro-F1 (26.6) and example-F1 (24.4) scores, indicating stronger generalization than prior methods.
Consistent with the MIMIC-CXR results (Table \ref{tab:single-image-comparison-MIMIC}), RA-RRG yielded lower NLG metric scores.
While comparable BLEU-1 scores suggested preservation of key content, ROUGE-L and BLEU-4 were lower, reflecting their reliance on exact lexical overlap.
The limited correlation between NLG metrics and clinical quality is further examined in Section \ref{subsec:nlg}.

Comparing CheXbert results across datasets revealed a substantial drop in Macro-F1 from 41.7 on MIMIC-CXR to 26.6 on IU X-Ray.
This discrepancy was consistent with prior observations that CheXpert labels are tailored to MIMIC-CXR and may be less suitable for IU X-Ray \cite{chexpert, r2gen}.
Moreover, since the RadGraph model used for key phrase extraction was trained on MIMIC-CXR and CheXpert data, reduced generalization on the held-out dataset was observed.
For the benefit of future research,
Table \ref{tab:single-image-comparison-IU} reports all clinical efficacy metrics for RA-RRG, including those not provided by \citet{promptmrg}.

\subsection{Multi-View RRG}\label{subsec:multi-view-rrg}

\begin{table*}[t]
\caption{
Results of multi-view RRG evaluation on the MIMIC-CXR test set.
The test setting follows MAIRA-2.
}
\centering
\small
\setlength{\tabcolsep}{3pt}
\renewcommand{\arraystretch}{1.12}
\label{tab:multi-image-comparison-radgraph}

\begin{adjustbox}{max width=\textwidth}
\begin{tabular}{l|ccc|c|ccc}
\Xhline{1pt}
\multirow{2}{*}{Model}
& \multicolumn{3}{c|}{CheXbert}
& \multirow{2}{*}{\makecell{RadGraph \\ F1}}
& \multicolumn{3}{c}{NLG Metrics} \\
\cline{2-4} \cline{6-8}
 & micro-F1 & Macro-F1 & example-F1
 & & ROUGE-L & BLEU-1 & BLEU-4 \\
\hline\hline

Med-PaLM M 84B \textsubscript{Zero-shot}
 & 50.5 & 37.8 & - & 28.3 & 28.7 & 34.6 & 12.4 \\

MAIRA-2 \textsubscript{Infer: No Prior No Comp}
 & - & 35.8 & - & - & 27.3 & - & - \\

MAIRA-2 \textsubscript{Train: No Prior No Comp}
 & - & 39.3 & - & - & 33.9 & - & - \\

\cline{1-8}
RA-RRG
 & \textbf{60.6} & \textbf{42.2} & \textbf{54.3}
 & 25.8 & 24.2 & 34.1 & 7.0 \\
\Xhline{1pt}
\end{tabular}
\end{adjustbox}
\end{table*}

Table \ref{tab:multi-image-comparison-radgraph} presents multi-view performance results on the MIMIC-CXR test set. 
We evaluated RA-RRG in a two-view setting alongside two comparison models.
Med-PaLM M 84B, a single-view model, reported zero-shot performance in a two-view setting, although the exact test configuration was unspecified.
MAIRA-2,
designed for multi-study inputs, was evaluated using its reported ablations on 2,181 studies with prior information.

Med-PaLM M 84B achieved reasonable RadGraph F1 scores, 
likely due to its large model size, 
but showed substantially lower CheXbert performance than RA-RRG.
MAIRA-2 performed poorly without prior information (Macro-F1 35.8),
and even the trained version (39.3) fell below our RA-RRG, which achieved 42.2. 
Despite being trained only on single-image retrieval, RA-RRG showed effective generalization 
to multi-view inputs.

\subsection{Hallucination Analysis}
\label{subsec:hallucination}

\subsubsection{Comparative hallucination suppression}

\begin{table}[ht]
\caption{Frequency of comparative hallucination-related terms and the percentage of
model-generated reports containing each keyword.}
\centering
\small
\setlength{\tabcolsep}{2pt}
\renewcommand{\arraystretch}{1.1}
\label{tab:hallucination_keywords}

\begin{adjustbox}{max width=\columnwidth}
\begin{tabular}{l|ccc|ccc}
\Xhline{1pt}
\multirow{2}{*}{Keyword}
& \multicolumn{3}{c|}{Keyword Count $\downarrow$}
& \multicolumn{3}{c}{Report Inclusion Rate (\%) $\downarrow$} \\
\cline{2-7}
& PromptMRG & E1 & RA-RRG
& PromptMRG & E1 & RA-RRG \\
\hline\hline

Change        & 559 & 160 & \textbf{16} & 9.95 & 4.04 & \textbf{0.41} \\
Unchanged     & 1187 & 5992 & \textbf{1} & 24.44 & 73.46 & \textbf{0.03} \\
Prior         & 405 & 1141 & \textbf{38} & 7.70 & 24.31 & \textbf{0.98} \\
Stable        & 431 & 101 & \textbf{4} & 10.91 & 2.49 & \textbf{0.10} \\
Interval      & 359 & 69 & \textbf{0} & 6.53 & 1.79 & \textbf{0.00} \\
Previous      & 920 & 264 & \textbf{20} & 22.65 & 6.17 & \textbf{0.52} \\
Again         & 600 & 1280 & \textbf{0} & 12.93 & 27.94 & \textbf{0.00} \\
Increased     & \textbf{280} & 1125 & 869 & \textbf{6.95} & 23.43 & 19.10 \\
Improve       & \textbf{0} & \textbf{0} & \textbf{0} & \textbf{0.00} & \textbf{0.00} & \textbf{0.00} \\
Remain        & 104 & \textbf{25} & 76 & 2.62 & \textbf{0.65} & 1.97 \\
Worse         & 18 & 9 & \textbf{0} & 0.41 & 0.23 & \textbf{0.00} \\
Persistent    & 34 & 449 & \textbf{0} & 0.83 & 11.25 & \textbf{0.00} \\
Removal       & 49 & \textbf{6} & 8 & 1.11 & \textbf{0.16} & 0.21 \\
Similar       & 729 & 180 & \textbf{5} & 15.66 & 4.54 & \textbf{0.13} \\
Earlier       & 84 & \textbf{3} & \textbf{3} & 2.07 & \textbf{0.08} & \textbf{0.08} \\
Decreased     & 516 & 131 & \textbf{31} & 11.28 & 3.21 & \textbf{0.80} \\
Recurrence    & \textbf{0} & \textbf{0} & \textbf{0} & \textbf{0.00} & \textbf{0.00} & \textbf{0.00} \\
Redemonstrate & \textbf{0} & \textbf{0} & \textbf{0} & \textbf{0.00} & \textbf{0.00} & \textbf{0.00} \\
\Xhline{1pt}
\end{tabular}
\end{adjustbox}
\end{table}

\begin{figure}[h]
    \begin{center}
    \scalebox{1.0}{
        \includegraphics[width=\linewidth]{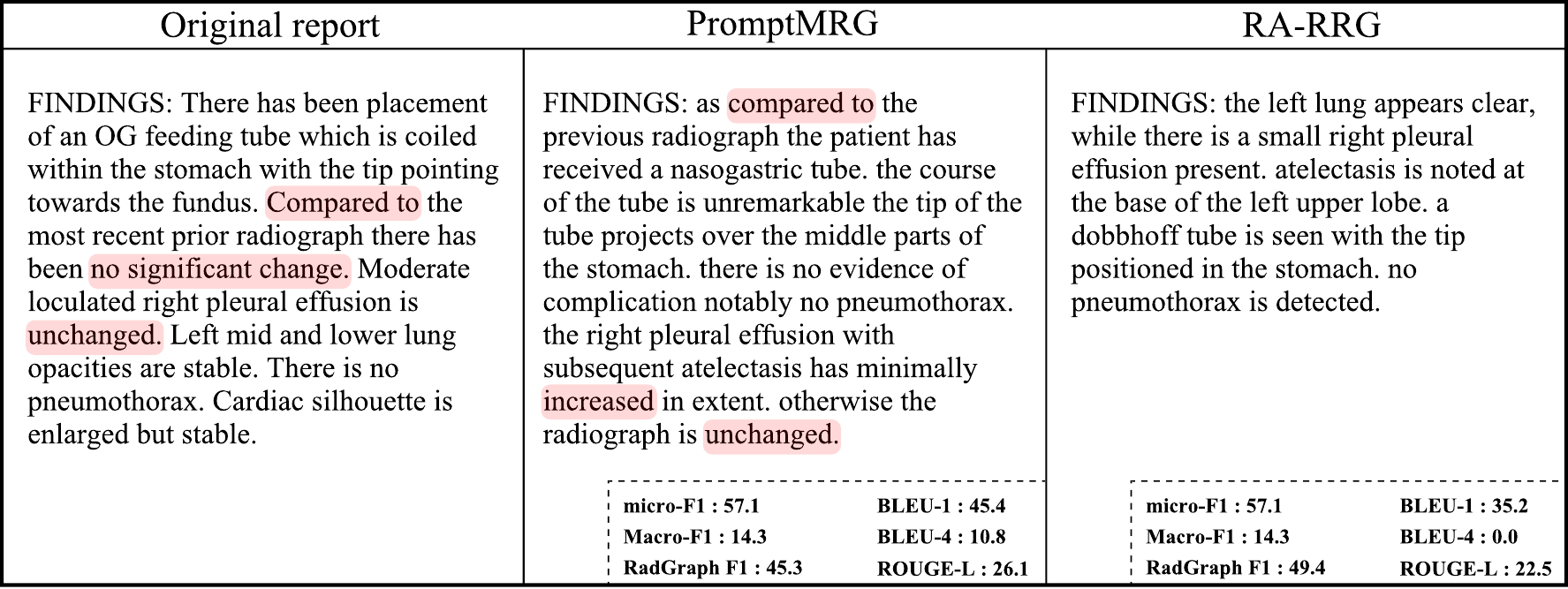}
    }
    \vspace{-0.5cm}
    \end{center}
    \caption{
    Example reports generated by PromptMRG \cite{promptmrg} and RA-RRG for a single MIMIC-CXR test image. Hallucinated expressions are highlighted in red, with clinical accuracy and NLG metrics reported.
    }
    \vspace{-0.3cm}
    \label{fig:hallucination}
\end{figure}

Table \ref{tab:hallucination_keywords} compares three models—PromptMRG, a sentence-level extraction ablation (E1), and RA-RRG—using two metrics: the frequency of comparative hallucination-related keywords and the percentage of reports containing them.
E1 segments reports at the sentence level rather than extracting key phrases,
with details provided in Appendix \ref{subsec:ablation}.
Unlike PromptMRG and E1, both 
trained on sentences referencing prior studies, RA-RRG exhibited lower keyword frequency and fewer affected reports.
With RadGraph and 
LLM-extracted key phrases,
RA-RRG effectively removed unnecessary comparisons and irrelevant content.
Figure \ref{fig:hallucination} illustrates a representative example: while PromptMRG generated comparative expressions such as `compared to' and `unchanged', RA-RRG produced an accurate, concise, and hallucination-free report.

In Table \ref{tab:hallucination_keywords},
most comparative hallucination-related keywords appeared in less than 1\% of the 
reports generated by RA-RRG,
except for the term `increased' which was found in 19.1\%.
While we directly adopted the term set from \citet{cxr-redone},
`increased' does not necessarily imply comparison with prior studies. 
For example, in the phrase `increased interstitial marking,' 
the term refers to a deviation from the normal state, 
which 
is identifiable from a single image.
Our analysis revealed that
32\% (278 out of 869) of `increased' 
appeared in this specific phrase.
To enable more rigorous comparative hallucination detection, 
the term set should be carefully curated.

\subsubsection{Entity- and relation-level analysis}

\begin{table}[t]
\caption{
Entity- and relation-level analysis comparing the sentence-level baseline (E1) and RA-RRG.
CheXbert scores are example-based, and RadGraph scores are computed at the entity and relation levels.
}
\centering
\small
\setlength{\tabcolsep}{3pt}
\renewcommand{\arraystretch}{1.12}
\label{tab:entity_relation}

\begin{adjustbox}{max width=\columnwidth}
\begin{tabular}{l|ccc|ccc}
\Xhline{1pt}
\multirow{2}{*}{Metric} & \multicolumn{3}{c|}{E1} & \multicolumn{3}{c}{RA-RRG} \\
\cline{2-7}
 & Precision & Recall & F1 & Precision & Recall & F1 \\
\hline\hline
CheXbert & \textbf{0.492} & 0.557 & 0.489 & 0.491 & \textbf{0.599} & \textbf{0.507} \\
RadGraph entity & 0.311 & 0.402 & 0.339 & \textbf{0.334} & \textbf{0.416} & \textbf{0.360} \\
RadGraph relation & 0.133 & 0.183 & 0.147 & \textbf{0.160} & \textbf{0.209} & \textbf{0.173} \\
\Xhline{1pt}
\end{tabular}
\end{adjustbox}
\end{table}

While keyword-based analysis captures comparative hallucinations, it 
fails to address 
fabricated clinical findings.
To evaluate object-level hallucinations, we compare E1 and RA-RRG using entity- and relation-level RadGraph metrics alongside CheXbert example-based precision, recall, and F1 (Table~\ref{tab:entity_relation}).
RA-RRG improved CheXbert example-based recall (0.557 $\to$ 0.599) while maintaining comparable precision (0.492 $\to$ 0.491), leading to a higher F1 (0.489 $\to$ 0.507).
At both the RadGraph entity and relation levels, RA-RRG improved precision, recall, and F1, 
indicating that RA-RRG not only suppresses comparative expressions but also reduces false-positive fabrications while improving factual coverage.

\subsection{Retrieval Error Analysis}
\label{subsec:retrieval-error}

\begin{table}[h]
\caption{
Comparison of micro-averaged CheXbert and RadGraph F1 before and after LLM-based generation.
}
\centering
\small
\setlength{\tabcolsep}{3pt}
\renewcommand{\arraystretch}{1.12}
\label{tab:stage_comparison}

\begin{adjustbox}{max width=\columnwidth}
\begin{tabular}{l|c|ccc|c}
\Xhline{1pt}
\multirow{2}{*}{Stage} & \multicolumn{4}{c|}{CheXbert (micro-averaged)} & \multirow{2}{*}{\makecell{RadGraph \\ F1}} \\
\cline{2-5}
 & F1 & Recall & Precision & Specificity & \\
\hline\hline
Retrieval-only & 0.588 & 0.681 & 0.517 & 0.878 & 0.257 \\
After Generation & 0.585 & 0.671 & 0.519 & 0.881 & 0.267 \\
\Xhline{1pt}
\end{tabular}
\end{adjustbox}
\end{table}

To analyze how retrieval errors propagate through the pipeline, we compare micro-averaged CheXbert performance and RadGraph F1 before and after LLM-based generation (Table~\ref{tab:stage_comparison}).
In the retrieval-only setting, retrieved key phrases are 
simply concatenated with period delimiters
without LLM generation.
After generation, 
recall decreased slightly, whereas precision and specificity increased marginally.
The higher RadGraph F1 scores after generation indicate that the reports generated by the LLM better preserve entity-relation structures.
Overall, LLM-based generation mildly pruned 
over-predicted findings rather than introducing additional fabrications, 
although retrieval errors largely propagated to the final output.
A label-level analysis is provided in Appendix \ref{subsec:label-level-retrieval}.

\subsection{Computational Cost}
\label{subsec:computational-cost}

\begin{table}[t]
\caption{
Comparison of training computational cost.
GPU-hours are computed as the number of GPUs multiplied by training time.
}
\centering
\small
\setlength{\tabcolsep}{4pt}
\renewcommand{\arraystretch}{1.12}
\label{tab:training_cost}

\begin{tabular}{l|c|c|c}
\Xhline{1pt}
Method & Hardware & Time (hours) & Total GPU-hours \\
\hline\hline
M4CXR & 2$\times$H100 & 108 & 208 \\
LLaVA-Rad & 8$\times$A100 & 28 & 224 \\
RA-RRG & 1$\times$H100 & 18 & 18 \\
\Xhline{1pt}
\end{tabular}
\end{table}

Table~\ref{tab:training_cost} compares the training computational cost of RA-RRG with comparable MLLM baselines.
RA-RRG required 
only 18 GPU-hours on a single H100, compared to 208 GPU-hours for M4CXR and 224 GPU-hours for LLaVA-Rad.
While H100-hours and A100-hours are not directly interchangeable, the comparison highlights that RA-RRG requires substantially fewer GPU-hours and fewer devices than multimodal LLM training pipelines.

\section{Discussion}

\subsection{Limitations of NLG Metrics for RRG} 
\label{subsec:nlg}

RA-RRG consistently reports lower NLG scores than other models (Sections \ref{subsec:single-view}, \ref{subsec:multi-view-rrg}) because it focuses on key phrase extraction, omitting irrelevant details such as view positions or comparisons with prior images, which reduces lexical overlap while preserving clinical relevance.
We argue that clinical accuracy should be prioritized over NLG metrics 
for RRG.
NLG metrics rely on surface-level similarity and can be misleading. 
For example, 
given a ground truth report `Bilateral pleural effusions are present',
a generated report `Bilateral pleural effusions are not present' receives high NLG scores 
(ROUGE-L: 0.9242, BLEU-1: 0.8333, BLEU-4: 0.5373)
despite contradicting the reference text.

RA-RRG rephrases retrieved key phrases using an LLM, which can lower NLG scores despite clinical correctness.
As shown in Figure \ref{fig:hallucination}, RA-RRG achieved lower NLG scores than PromptMRG while matching its CheXbert scores, indicating that both models captured key findings and that the NLG gap mainly reflects structural variation rather than clinical errors.
Moreover, its higher RadGraph F1 score suggests better preservation of clinical relations despite rephrased text.

\begin{figure*}[ht!]
    \begin{center}
    \scalebox{0.95}{
        \includegraphics[width=\linewidth]{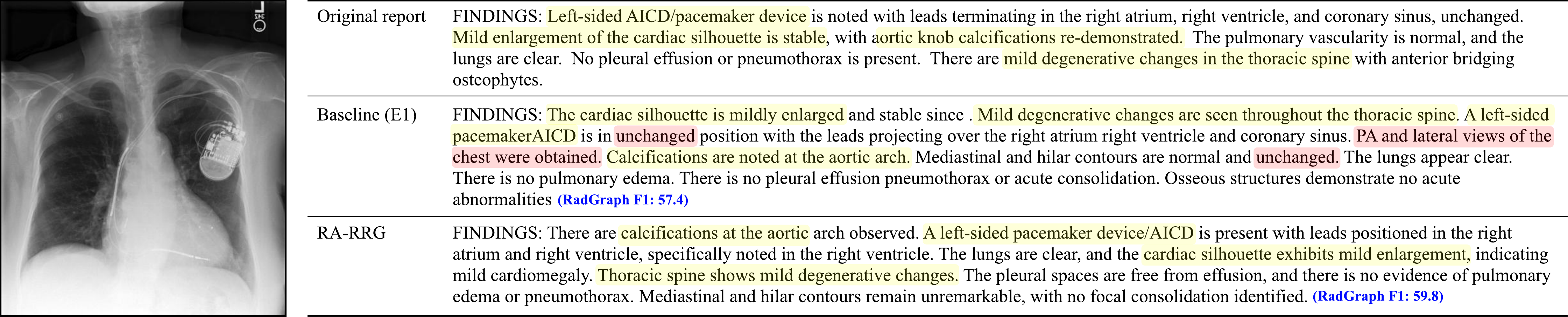}
    }
    \vspace{-0.5cm}
    \end{center}
    \caption{
    Example of single-view RRG. The baseline is sentence-level ablation (E1) from Table \ref{tab:Ablation} in Appendix. 
    Positive findings are highlighted in yellow, and hallucinations are marked in red. 
    }
    \vspace{-0.3cm}
    \label{fig:single-qualitative}
\end{figure*}

\begin{figure*}[ht!]
    \begin{center}
    \scalebox{0.95}{
        \includegraphics[width=\linewidth]{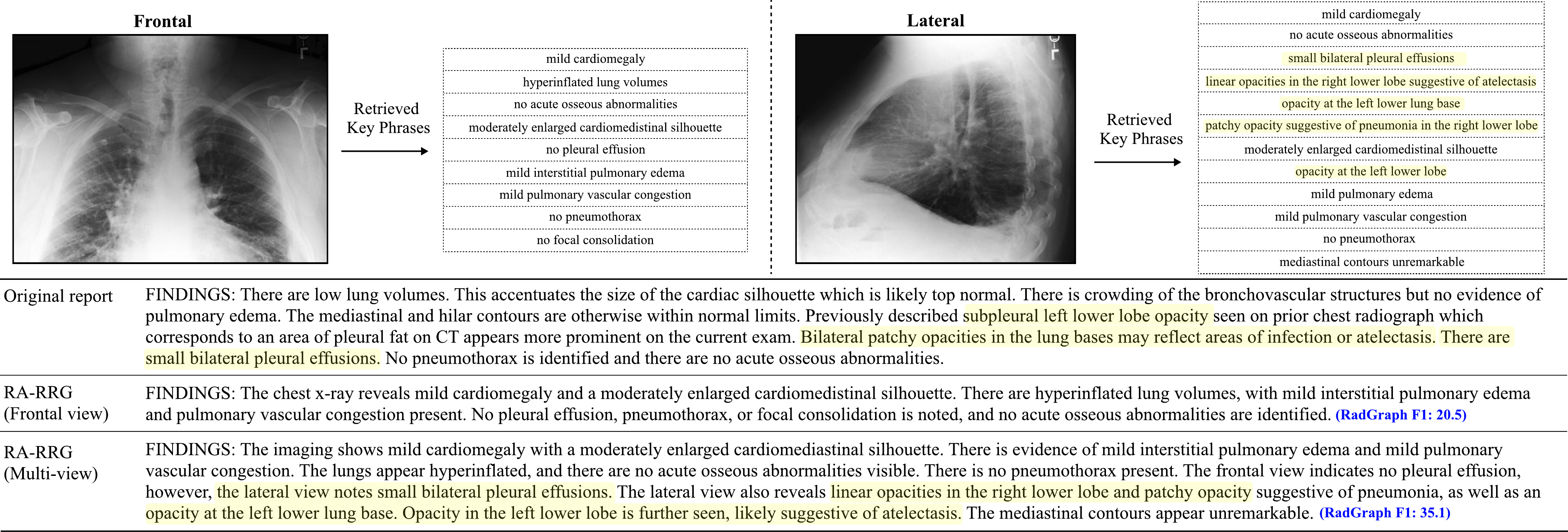}
    }
    \vspace{-0.5cm}
    \end{center}
    \caption{
    Example of multi-view RRG. At the top are the frontal and lateral images with their predicted key phrases. 
    Below the original report, two radiology reports are generated: 1) using only the frontal view, and 2) using both the frontal and lateral views (multi-view).
    Content present in the original report but visible only in the lateral view is highlighted in yellow. 
    }
    \vspace{-0.3cm}
    \label{fig:multi-qualitative}
\end{figure*}

\subsection{Qualitative Analysis}

Figure \ref{fig:single-qualitative} presents an example of single-view RRG. 
For comparison, we also include the output of E1, a sentence-level extraction baseline described in Section \ref{subsec:hallucination}.
Both E1 and RA-RRG accurately predicted
positive findings highlighted in yellow,
such as enlarged cardiac silhouette and calcification.
However, 
E1 exhibited hallucinations by generating comparative expressions such as ``unchanged" 
and referencing 
multiple views,
despite being given only a single frontal image.
In contrast, 
RA-RRG avoided such hallucinations 
and produced a concise, focused report consistent with the input.

Figure \ref{fig:multi-qualitative} shows an example of multi-view RRG, comparing reports generated using only the frontal-view key phrases with those incorporating both frontal and lateral views. 
When using only the frontal view, the model missed 
pleural effusion and opacity-related findings.
By incorporating lateral-view key phrases, the model correctly identified bilateral pleural effusion, opacity, and suspected atelectasis.
Although the multi-view report 
missed suspected pneumonia,
it showed improved diagnostic performance over 
the frontal-only report. 

\section{Conclusion}

In this study, we introduced RA-RRG, a retrieval-augmented framework for RRG that leveraged LLMs.
By extracting clinically essential key phrases and retrieving image-consistent phrases, RA-RRG effectively suppressed hallucinations and generated clinically faithful reports.
Experimental results demonstrated that RA-RRG achieved strong performance on standard clinical metrics, remaining competitive with fine-tuned multimodal LLMs without requiring any LLM fine-tuning, while using substantially fewer computational resources.
Analysis at the entity and relation level further confirmed that RA-RRG reduces hallucinations beyond comparative expressions.
The proposed framework naturally generalized to multi-view RRG
by aggregating phrases retrieved from multiple images.
Accordingly, this retrieval-based paradigm 
enabled the extension of
RRG to broader clinical settings without additional model training.

\section*{Limitations}

Despite the robust performance of RA-RRG, the proposed framework has several limitations.
First, RA-RRG relies on LLMs for key phrase extraction, and the quality of the extracted phrases can be sensitive to the capability of the underlying LLM.
Our ablation analysis in Appendix \ref{subsec:ablation} indicates that
the RRG stage itself does not heavily depend on LLM performance; however, errors or omissions in key phrase extraction may propagate to subsequent retrieval and report generation stages.
In addition, retrieval errors cannot be corrected during report generation, since the LLM conditions its output on the retrieved key phrases.
While this design helps suppress hallucinations, it also limits the model’s ability to recover from retrieval-stage errors.
Future work could explore lightweight refinement mechanisms to mitigate error propagation without reintroducing hallucinations.

Our evaluation is subject to limitations regarding
coverage and clinical assessment.
The phrase-level vector database is constructed from key phrases extracted solely from the training set, which may limit the system’s ability to handle out-of-vocabulary findings.
In addition, our evaluation lacks 
human assessment by radiologists; while we report standard clinical and automatic metrics, expert evaluation 
remains necessary to rigorously
assess clinical correctness, usefulness, and readability in real-world settings.
Incorporating broader phrase coverage and human evaluation will be essential for validating the framework in real clinical practice.

\section*{Acknowledgements}

This work was supported by the Technology Innovation Program (RS-2025-02221011, Development of Medical-Specialized Multimodal Hyperscale Generative AI Technology for Global Integration) funded by the Ministry of Trade Industry \& Energy (MOTIE, South Korea),
and by the faculty research fund of Sejong University in 2026.

\bibliography{custom}

@article{maira1,
  title={MAIRA-1: A specialised large multimodal model for radiology report generation}, 
  author={Stephanie L. Hyland and Shruthi Bannur and Kenza Bouzid and Daniel C. Castro and Mercy Ranjit and Anton Schwaighofer and Fernando Pérez-García and Valentina Salvatelli and Shaury Srivastav and Anja Thieme and Noel Codella and Matthew P. Lungren and Maria Teodora Wetscherek and Ozan Oktay and Javier Alvarez-Valle},
  journal={arXiv preprint arXiv:2311.13668},
  year={2023}
}

@article{maira2,
  title={MAIRA-2: Grounded Radiology Report Generation},
  author={Bannur, Shruthi and Bouzid, Kenza and Castro, Daniel C and Schwaighofer, Anton and Bond-Taylor, Sam and Ilse, Maximilian and P{\'e}rez-Garc{\'\i}a, Fernando and Salvatelli, Valentina and Sharma, Harshita and Meissen, Felix and others},
  journal={arXiv preprint arXiv:2406.04449},
  year={2024}
}

@article{promptmrg, 
    title={PromptMRG: Diagnosis-Driven Prompts for Medical Report Generation}, 
    volume={38}, 
    url={https://ojs.aaai.org/index.php/AAAI/article/viTowardsew/28038}, 
    DOI={10.1609/aaai.v38i3.28038}, 
    number={3}, 
    journal={Proceedings of the AAAI Conference on Artificial Intelligence}, 
    author={Jin, Haibo and Che, Haoxuan and Lin, Yi and Chen, Hao}, 
    year={2024}, 
    month={Mar.}, 
    pages={2607-2615} 
}

@article{chexagent,
  title={CheXagent: Towards a Foundation Model for Chest X-Ray Interpretation}, 
  author={Zhihong Chen and Maya Varma and Jean-Benoit Delbrouck and Magdalini Paschali and Louis Blankemeier and Dave Van Veen and Jeya Maria Jose Valanarasu and Alaa Youssef and Joseph Paul Cohen and Eduardo Pontes Reis and Emily B. Tsai and Andrew Johnston and Cameron Olsen and Tanishq Mathew Abraham and Sergios Gatidis and Akshay S. Chaudhari and Curtis Langlotz},
  journal={arXiv preprint arXiv:2401.12208},
  year={2024}
}

@Article{llava-rad,
author={Juan Manuel Zambrano Chaves
and Shih-Cheng Huang
and Yanbo Xu
and Hanwen Xu
and Naoto Usuyama
and Sheng Zhang
and Fei Wang
and Yujia Xie
and Mahmoud Khademi
and Ziyi Yang
and Hany Awadalla
and Julia Gong
and Houdong Hu
and Jianwei Yang
and Chunyuan Li
and Jianfeng Gao
and Yu Gu
and Cliff Wong
and Mu Wei
and Tristan Naumann
and Muhao Chen
and Matthew P. Lungren
and Akshay Chaudhari
and Serena Yeung-Levy
and Curtis P. Langlotz
and Sheng Wang
and Hoifung Poon},
title={A clinically accessible small multimodal radiology model and evaluation metric for chest X-ray findings},
journal={Nature Communications},
year={2025},
month={Apr},
day={01},
volume={16},
number={1},
pages={3108},
issn={2041-1723},
doi={10.1038/s41467-025-58344-x},
url={https://doi.org/10.1038/s41467-025-58344-x}
}

@article{medpalm-m,
  title={Towards generalist biomedical AI},
  author={Tu, Tao and Azizi, Shekoofeh and Driess, Danny and Schaekermann, Mike and Amin, Mohamed and Chang, Pi-Chuan and Carroll, Andrew and Lau, Charles and Tanno, Ryutaro and Ktena, Ira and others},
  journal={NEJM AI},
  volume={1},
  number={3},
  pages={AIoa2300138},
  year={2024},
  publisher={Massachusetts Medical Society}
}

@article{medgemini2d,
  title={Advancing multimodal medical capabilities of Gemini},
  author={Yang, Lin and Xu, Shawn and Sellergren, Andrew and Kohlberger, Timo and Zhou, Yuchen and Ktena, Ira and Kiraly, Atilla and Ahmed, Faruk and Hormozdiari, Farhad and Jaroensri, Tiam and others},
  journal={arXiv preprint arXiv:2405.03162},
  year={2024}
}

@article{mimic-cxr,
  title={MIMIC-CXR, a de-identified publicly available database of chest radiographs with free-text reports},
  author={Johnson, Alistair EW and Pollard, Tom J and Berkowitz, Seth J and Greenbaum, Nathaniel R and Lungren, Matthew P and Deng, Chih-ying and Mark, Roger G and Horng, Steven},
  journal={Scientific data},
  volume={6},
  number={1},
  pages={317},
  year={2019},
  publisher={Nature Publishing Group UK London}
}

@article{mimic-cxr-jpg,
  title={MIMIC-CXR-JPG, a large publicly available database of labeled chest radiographs},
  author={Johnson, Alistair EW and Pollard, Tom J and Greenbaum, Nathaniel R and Lungren, Matthew P and Deng, Chih-ying and Peng, Yifan and Lu, Zhiyong and Mark, Roger G and Berkowitz, Seth J and Horng, Steven},
  journal={arXiv preprint arXiv:1901.07042},
  year={2019}
}

@inproceedings{chexbert,
  title={CheXbert: Combining automatic labelers and expert annotations for accurate radiology report labeling using BERT},
  author={Smit, Akshay and Jain, Saahil and Rajpurkar, Pranav and Pareek, Anuj and Ng, Andrew Y and Lungren, Matthew P},
  booktitle={EMNLP 2020-2020 Conference on Empirical Methods in Natural Language Processing, Proceedings of the Conference},
  pages={1500--1519},
  year={2020}
}

@inproceedings{bleu,
  title={Bleu: a method for automatic evaluation of machine translation},
  author={Papineni, Kishore and Roukos, Salim and Ward, Todd and Zhu, Wei-Jing},
  booktitle={Proceedings of the 40th annual meeting of the Association for Computational Linguistics},
  pages={311--318},
  year={2002}
}

@inproceedings{rouge,
  title={Rouge: A package for automatic evaluation of summaries},
  author={Lin, Chin-Yew},
  booktitle={Text summarization branches out},
  pages={74--81},
  year={2004}
}

@inproceedings{radgraph,
 author = {Jain, Saahil and Agrawal, Ashwin and Saporta, Adriel and Truong, Steven and Duong, Du Nguyen Duong Nguyen and Bui, Tan and Chambon, Pierre and Zhang, Yuhao and Lungren, Matthew and Ng, Andrew and Langlotz, Curtis and Rajpurkar, Pranav and Rajpurkar, Pranav},
 booktitle = {Proceedings of the Neural Information Processing Systems Track on Datasets and Benchmarks},
 editor = {J. Vanschoren and S. Yeung},
 pages = {},
 title = {RadGraph: Extracting Clinical Entities and Relations from Radiology Reports},
 url = {https://datasets-benchmarks-proceedings.neurips.cc/paper_files/paper/2021/file/c8ffe9a587b126f152ed3d89a146b445-Paper-round1.pdf},
 volume = {1},
 year = {2021}
}

@article{radgraph-f1,
  title={Evaluating progress in automatic chest x-ray radiology report generation},
  author={Yu, Feiyang and Endo, Mark and Krishnan, Rayan and Pan, Ian and Tsai, Andy and Reis, Eduardo Pontes and Fonseca, Eduardo Kaiser Ururahy Nunes and Lee, Henrique Min Ho and Abad, Zahra Shakeri Hossein and Ng, Andrew Y and others},
  journal={Patterns},
  volume={4},
  number={9},
  year={2023},
  publisher={Elsevier}
}

@inproceedings{cxr-repair,
  title={Retrieval-based chest x-ray report generation using a pre-trained contrastive language-image model},
  author={Endo, Mark and Krishnan, Rayan and Krishna, Viswesh and Ng, Andrew Y and Rajpurkar, Pranav},
  booktitle={Machine Learning for Health},
  pages={209--219},
  year={2021},
  organization={PMLR}
}

@InProceedings{cxr-redone,
  title = 	 {Improving Radiology Report Generation Systems by Removing Hallucinated References to Non-existent Priors},
  author =       {Ramesh, Vignav and Chi, Nathan A. and Rajpurkar, Pranav},
  booktitle = 	 {Proceedings of the 2nd Machine Learning for Health symposium},
  pages = 	 {456--473},
  year = 	 {2022},
  editor = 	 {Parziale, Antonio and Agrawal, Monica and Joshi, Shalmali and Chen, Irene Y. and Tang, Shengpu and Oala, Luis and Subbaswamy, Adarsh},
  volume = 	 {193},
  series = 	 {Proceedings of Machine Learning Research},
  month = 	 {28 Nov},
  publisher =    {PMLR},
  url = 	 {https://proceedings.mlr.press/v193/ramesh22a.html},
}

@inproceedings{cxr-rag,
  title={Retrieval augmented chest x-ray report generation using openai gpt models},
  author={Ranjit, Mercy and Ganapathy, Gopinath and Manuel, Ranjit and Ganu, Tanuja},
  booktitle={Machine Learning for Healthcare Conference},
  pages={650--666},
  year={2023},
  organization={PMLR}
}

@article{iu-xray,
  author = {Demner-Fushman, Dina and Kohli, Marc D and Rosenman, Marc B and Shooshan, Steven E and Rodriguez, Louis and Antani, Sameer and Thoma, George R and McDonald, Clement J},
  title = {Preparing a collection of radiology examinations for distribution and retrieval},
  journal = {Journal of the American Medical Informatics Association},
  volume = {23},
  number = {2},
  pages = {304--310},
  year = {2016},
  month = {Mar},
  doi = {10.1093/jamia/ocv080},
  pmid = {26133894},
  pmcid = {PMC5009925}
}

@inproceedings{style-aware-rrg,
  title={Style-Aware Radiology Report Generation with RadGraph and Few-Shot Prompting},
  author={Yan, Benjamin and Liu, Ruochen and Kuo, David and Adithan, Subathra and Reis, Eduardo and Kwak, Stephen and Venugopal, Vasantha and O’Connell, Chloe and Saenz, Agustina and Rajpurkar, Pranav and others},
  booktitle={Findings of the Association for Computational Linguistics: EMNLP 2023},
  pages={14676--14688},
  year={2023}
}

@inproceedings{fact-aware-mrag,
    title = "Fact-Aware Multimodal Retrieval Augmentation for Accurate Medical Radiology Report Generation",
    author = "Sun, Liwen  and
      Zhao, James Jialun  and
      Han, Wenjing  and
      Xiong, Chenyan",
    editor = "Chiruzzo, Luis  and
      Ritter, Alan  and
      Wang, Lu",
    booktitle = "Proceedings of the 2025 Conference of the Nations of the Americas Chapter of the Association for Computational Linguistics: Human Language Technologies (Volume 1: Long Papers)",
    month = apr,
    year = "2025",
    address = "Albuquerque, New Mexico",
    publisher = "Association for Computational Linguistics",
    url = "https://aclanthology.org/2025.naacl-long.28/",
    doi = "10.18653/v1/2025.naacl-long.28",
    pages = "643--655",
    ISBN = "979-8-89176-189-6"
}

@article{rag,
  title={Retrieval-augmented generation for knowledge-intensive nlp tasks},
  author={Lewis, Patrick and Perez, Ethan and Piktus, Aleksandra and Petroni, Fabio and Karpukhin, Vladimir and Goyal, Naman and K{\"u}ttler, Heinrich and Lewis, Mike and Yih, Wen-tau and Rockt{\"a}schel, Tim and others},
  journal={Advances in Neural Information Processing Systems},
  volume={33},
  pages={9459--9474},
  year={2020}
}

@article{rag-survey,
  title={Retrieval-Augmented Generation for Large Language Models: A Survey}, 
  author={Yunfan Gao and Yun Xiong and Xinyu Gao and Kangxiang Jia and Jinliu Pan and Yuxi Bi and Yi Dai and Jiawei Sun and Meng Wang and Haofen Wang},
  journal={arXiv preprint arXiv:2312.10997},
  year={2023}
}

@article{hallucination,
author = {Huang, Lei and Yu, Weijiang and Ma, Weitao and Zhong, Weihong and Feng, Zhangyin and Wang, Haotian and Chen, Qianglong and Peng, Weihua and Feng, Xiaocheng and Qin, Bing and Liu, Ting},
title = {A Survey on Hallucination in Large Language Models: Principles, Taxonomy, Challenges, and Open Questions},
year = {2025},
issue_date = {March 2025},
publisher = {Association for Computing Machinery},
address = {New York, NY, USA},
volume = {43},
number = {2},
issn = {1046-8188},
url = {https://doi.org/10.1145/3703155},
doi = {10.1145/3703155},
journal = {ACM Trans. Inf. Syst.},
month = jan,
articleno = {42},
numpages = {55}
}

@inproceedings{evcap,
  title={EVCap: Retrieval-Augmented Image Captioning with External Visual-Name Memory for Open-World Comprehension},
  author={Li, Jiaxuan and Vo, Duc Minh and Sugimoto, Akihiro and Nakayama, Hideki},
  booktitle={Proceedings of the IEEE/CVF Conference on Computer Vision and Pattern Recognition},
  pages={13733--13742},
  year={2024}
}

@inproceedings{ravideocap,
  title={Retrieval-augmented egocentric video captioning},
  author={Xu, Jilan and Huang, Yifei and Hou, Junlin and Chen, Guo and Zhang, Yuejie and Feng, Rui and Xie, Weidi},
  booktitle={Proceedings of the IEEE/CVF Conference on Computer Vision and Pattern Recognition},
  pages={13525--13536},
  year={2024}
}

@inproceedings{smallcap,
  title={Smallcap: lightweight image captioning prompted with retrieval augmentation},
  author={Ramos, Rita and Martins, Bruno and Elliott, Desmond and Kementchedjhieva, Yova},
  booktitle={Proceedings of the IEEE/CVF Conference on Computer Vision and Pattern Recognition},
  pages={2840--2849},
  year={2023}
}

@inproceedings{sarto2022retrieval,
  title={Retrieval-augmented transformer for image captioning},
  author={Sarto, Sara and Cornia, Marcella and Baraldi, Lorenzo and Cucchiara, Rita},
  booktitle={Proceedings of the 19th international conference on content-based multimedia indexing},
  pages={1--7},
  year={2022}
}

@inproceedings{
reimagen,
title={Re-Imagen: Retrieval-Augmented Text-to-Image Generator},
author={Wenhu Chen and Hexiang Hu and Chitwan Saharia and William W. Cohen},
booktitle={The Eleventh International Conference on Learning Representations },
year={2023},
url={https://openreview.net/forum?id=XSEBx0iSjFQ}
}

@inproceedings{ramlm,
  title={Retrieval-Augmented Multimodal Language Modeling},
  author={Yasunaga, Michihiro and Aghajanyan, Armen and Shi, Weijia and James, Richard and Leskovec, Jure and Liang, Percy and Lewis, Mike and Zettlemoyer, Luke and Yih, Wen-Tau},
  booktitle={International Conference on Machine Learning},
  pages={39755--39769},
  year={2023},
  organization={PMLR}
}

@ARTICLE{m4cxr,
  author={Park, Jonggwon and Kim, Soobum and Yoon, Byungmu and Hyun, Jihun and Choi, Kyoyun},
  journal={IEEE Transactions on Neural Networks and Learning Systems}, 
  title={M4CXR: Exploring Multitask Potentials of Multimodal Large Language Models for Chest X-Ray Interpretation}, 
  year={2025},
  volume={36},
  number={10},
  pages={17841-17855},
  doi={10.1109/TNNLS.2025.3587687}}

@inproceedings{hipporag,
title={Hippo{RAG}: Neurobiologically Inspired Long-Term Memory for Large Language Models},
author={Bernal Jimenez Gutierrez and Yiheng Shu and Yu Gu and Michihiro Yasunaga and Yu Su},
booktitle={The Thirty-eighth Annual Conference on Neural Information Processing Systems},
year={2024},
url={https://openreview.net/forum?id=hkujvAPVsg}
}

@InProceedings{transq,
author="Kong, Ming
and Huang, Zhengxing
and Kuang, Kun
and Zhu, Qiang
and Wu, Fei",
editor="Wang, Linwei
and Dou, Qi
and Fletcher, P. Thomas
and Speidel, Stefanie
and Li, Shuo",
title="TranSQ: Transformer-Based Semantic Query for Medical Report Generation",
booktitle="Medical Image Computing and Computer Assisted Intervention -- MICCAI 2022",
year="2022",
publisher="Springer Nature Switzerland",
address="Cham",
pages="610--620",
}

@article{teaser,
  title={Topicwise Separable Sentence Retrieval for Medical Report Generation},
  author={Zhao, Junting and Zhou, Yang and Chen, Zhihao and Fu, Huazhu and Wan, Liang},
  journal={IEEE Transactions on Medical Imaging},
  year={2024},
  publisher={IEEE}
}

@article{llama3,
  title={The llama 3 herd of models},
  author={Dubey, Abhimanyu and Jauhri, Abhinav and Pandey, Abhinav and Kadian, Abhishek and Al-Dahle, Ahmad and Letman, Aiesha and Mathur, Akhil and Schelten, Alan and Yang, Amy and Fan, Angela and others},
  journal={arXiv preprint arXiv:2407.21783},
  year={2024}
}

@inproceedings{detr,
  title={End-to-end object detection with transformers},
  author={Carion, Nicolas and Massa, Francisco and Synnaeve, Gabriel and Usunier, Nicolas and Kirillov, Alexander and Zagoruyko, Sergey},
  booktitle={European conference on computer vision},
  pages={213--229},
  year={2020},
  organization={Springer}
}

@inproceedings{clip,
  title={Learning transferable visual models from natural language supervision},
  author={Radford, Alec and Kim, Jong Wook and Hallacy, Chris and Ramesh, Aditya and Goh, Gabriel and Agarwal, Sandhini and Sastry, Girish and Askell, Amanda and Mishkin, Pamela and Clark, Jack and others},
  booktitle={International conference on machine learning},
  pages={8748--8763},
  year={2021},
  organization={PMLR}
}

@article{dinov2,
title={{DINO}v2: Learning Robust Visual Features without Supervision},
author={Maxime Oquab and Timoth{\'e}e Darcet and Th{\'e}o Moutakanni and Huy V. Vo and Marc Szafraniec and Vasil Khalidov and Pierre Fernandez and Daniel HAZIZA and Francisco Massa and Alaaeldin El-Nouby and Mido Assran and Nicolas Ballas and Wojciech Galuba and Russell Howes and Po-Yao Huang and Shang-Wen Li and Ishan Misra and Michael Rabbat and Vasu Sharma and Gabriel Synnaeve and Hu Xu and Herve Jegou and Julien Mairal and Patrick Labatut and Armand Joulin and Piotr Bojanowski},
journal={Transactions on Machine Learning Research},
issn={2835-8856},
year={2024},
url={https://openreview.net/forum?id=a68SUt6zFt},
note={}
}

@inproceedings{
eagle,
title={Eagle: Exploring The Design Space for Multimodal {LLM}s with Mixture of Encoders},
author={Min Shi and Fuxiao Liu and Shihao Wang and Shijia Liao and Subhashree Radhakrishnan and Yilin Zhao and De-An Huang and Hongxu Yin and Karan Sapra and Yaser Yacoob and Humphrey Shi and Bryan Catanzaro and Andrew Tao and Jan Kautz and Zhiding Yu and Guilin Liu},
booktitle={The Thirteenth International Conference on Learning Representations},
year={2025},
url={https://openreview.net/forum?id=Y2RW9EVwhT}
}

@inproceedings{neftune,
title={{NEFT}une: Noisy Embeddings Improve Instruction Finetuning},
author={Neel Jain and Ping-yeh Chiang and Yuxin Wen and John Kirchenbauer and Hong-Min Chu and Gowthami Somepalli and Brian R. Bartoldson and Bhavya Kailkhura and Avi Schwarzschild and Aniruddha Saha and Micah Goldblum and Jonas Geiping and Tom Goldstein},
booktitle={The Twelfth International Conference on Learning Representations},
year={2024},
url={https://openreview.net/forum?id=0bMmZ3fkCk}
}

@article{hungarian,
  title={The Hungarian method for the assignment problem},
  author={Kuhn, Harold W},
  journal={Naval research logistics quarterly},
  volume={2},
  number={1-2},
  pages={83--97},
  year={1955},
  publisher={Wiley Online Library}
}

@InProceedings{dbloss,
author="Wu, Tong
and Huang, Qingqiu
and Liu, Ziwei
and Wang, Yu
and Lin, Dahua",
editor="Vedaldi, Andrea
and Bischof, Horst
and Brox, Thomas
and Frahm, Jan-Michael",
title="Distribution-Balanced Loss for Multi-label Classification in Long-Tailed Datasets",
booktitle="Computer Vision -- ECCV 2020",
year="2020",
publisher="Springer International Publishing",
address="Cham",
pages="162--178",
isbn="978-3-030-58548-8"
}

@article{m2kt,
title = {Radiology report generation with a learned knowledge base and multi-modal alignment},
journal = {Medical Image Analysis},
volume = {86},
pages = {102798},
year = {2023},
issn = {1361-8415},
doi = {https://doi.org/10.1016/j.media.2023.102798},
url = {https://www.sciencedirect.com/science/article/pii/S1361841523000592},
author = {Shuxin Yang and Xian Wu and Shen Ge and Zhuozhao Zheng and S. Kevin Zhou and Li Xiao}
}

@article{cvt2dis.,
	title = {Improving chest {X}-ray report generation by leveraging warm starting},
	volume = {144},
	issn = {0933-3657},
	url = {https://www.sciencedirect.com/science/article/pii/S0933365723001471},
	doi = {https://doi.org/10.1016/j.artmed.2023.102633},
	journal = {Artificial Intelligence in Medicine},
	author = {Nicolson, Aaron and Dowling, Jason and Koopman, Bevan},
	year = {2023},
	pages = {102633},
}

@inproceedings{r2gen,
    title = "Generating Radiology Reports via Memory-driven Transformer",
    author = "Chen, Zhihong  and
      Song, Yan  and
      Chang, Tsung-Hui  and
      Wan, Xiang",
    editor = "Webber, Bonnie  and
      Cohn, Trevor  and
      He, Yulan  and
      Liu, Yang",
    booktitle = "Proceedings of the 2020 Conference on Empirical Methods in Natural Language Processing (EMNLP)",
    month = nov,
    year = "2020",
    address = "Online",
    publisher = "Association for Computational Linguistics",
    url = "https://aclanthology.org/2020.emnlp-main.112",
    doi = "10.18653/v1/2020.emnlp-main.112",
    pages = "1439--1449",
}

@inproceedings{rgrg,
   title={Interactive and Explainable Region-guided Radiology Report Generation},
   url={http://dx.doi.org/10.1109/CVPR52729.2023.00718},
   DOI={10.1109/cvpr52729.2023.00718},
   booktitle={2023 IEEE/CVF Conference on Computer Vision and Pattern Recognition (CVPR)},
   publisher={IEEE},
   author={Tanida, Tim and Müller, Philip and Kaissis, Georgios and Rueckert, Daniel},
   year={2023},
   month=jun, pages={7433–7442} }

@article{biomedclip,
author = {Sheng Zhang  and Yanbo Xu  and Naoto Usuyama  and Hanwen Xu  and Jaspreet Bagga  and Robert Tinn  and Sam Preston  and Rajesh Rao  and Mu Wei  and Naveen Valluri  and Cliff Wong  and Andrea Tupini  and Yu Wang  and Matt Mazzola  and Swadheen Shukla  and Lars Liden  and Jianfeng Gao  and Angela Crabtree  and Brian Piening  and Carlo Bifulco  and Matthew P. Lungren  and Tristan Naumann  and Sheng Wang  and Hoifung Poon },
title = {A Multimodal Biomedical Foundation Model Trained from Fifteen Million Image–Text Pairs},
journal = {NEJM AI},
volume = {2},
number = {1},
pages = {AIoa2400640},
year = {2025},
doi = {10.1056/AIoa2400640},

URL = {https://ai.nejm.org/doi/full/10.1056/AIoa2400640},
eprint = {https://ai.nejm.org/doi/pdf/10.1056/AIoa2400640}
}

@article{chexpertplus,
  title={CheXpert Plus: Hundreds of Thousands of Aligned Radiology Texts, Images and Patients},
  author={Chambon, Pierre and Delbrouck, Jean-Benoit and Sounack, Thomas and Huang, Shih-Cheng and Chen, Zhihong and Varma, Maya and Truong, Steven QH and Chuong, Chu The and Langlotz, Curtis P},
  journal={arXiv preprint arXiv:2405.19538},
  year={2024}
}

@Article{raddino,
author={P{\'e}rez-Garc{\'i}a, Fernando
and Sharma, Harshita
and Bond-Taylor, Sam
and Bouzid, Kenza
and Salvatelli, Valentina
and Ilse, Maximilian
and Bannur, Shruthi
and Castro, Daniel C.
and Schwaighofer, Anton
and Lungren, Matthew P.
and Wetscherek, Maria Teodora
and Codella, Noel
and Hyland, Stephanie L.
and Alvarez-Valle, Javier
and Oktay, Ozan},
title={Exploring scalable medical image encoders beyond text supervision},
journal={Nature Machine Intelligence},
year={2025},
month={Jan},
day={01},
volume={7},
number={1},
pages={119-130},
issn={2522-5839},
doi={10.1038/s42256-024-00965-w},
url={https://doi.org/10.1038/s42256-024-00965-w}
}

@inproceedings{sbert,
    title = "Sentence-{BERT}: Sentence Embeddings using {S}iamese {BERT}-Networks",
    author = "Reimers, Nils  and
      Gurevych, Iryna",
    editor = "Inui, Kentaro  and
      Jiang, Jing  and
      Ng, Vincent  and
      Wan, Xiaojun",
    booktitle = "Proceedings of the 2019 Conference on Empirical Methods in Natural Language Processing and the 9th International Joint Conference on Natural Language Processing (EMNLP-IJCNLP)",
    month = nov,
    year = "2019",
    address = "Hong Kong, China",
    publisher = "Association for Computational Linguistics",
    url = "https://aclanthology.org/D19-1410/",
    doi = "10.18653/v1/D19-1410",
    pages = "3982--3992"
}

@inproceedings{metransformer,
  title={Metransformer: Radiology report generation by transformer with multiple learnable expert tokens},
  author={Wang, Zhanyu and Liu, Lingqiao and Wang, Lei and Zhou, Luping},
  booktitle={Proceedings of the IEEE/CVF Conference on Computer Vision and Pattern Recognition},
  pages={11558--11567},
  year={2023}
}

@inproceedings{vllm,
  title={Efficient Memory Management for Large Language Model Serving with PagedAttention},
  author={Woosuk Kwon and Zhuohan Li and Siyuan Zhuang and Ying Sheng and Lianmin Zheng and Cody Hao Yu and Joseph E. Gonzalez and Hao Zhang and Ion Stoica},
  booktitle={Proceedings of the ACM SIGOPS 29th Symposium on Operating Systems Principles},
  year={2023}
}

@inproceedings{dcl,
  title={Dynamic graph enhanced contrastive learning for chest x-ray report generation},
  author={Li, Mingjie and Lin, Bingqian and Chen, Zicong and Lin, Haokun and Liang, Xiaodan and Chang, Xiaojun},
  booktitle={Proceedings of the IEEE/CVF Conference on Computer Vision and Pattern Recognition},
  pages={3334--3343},
  year={2023}
}

@inproceedings{chexpert,
author = {Irvin, Jeremy and Rajpurkar, Pranav and Ko, Michael and Yu, Yifan and Ciurea-Ilcus, Silviana and Chute, Chris and Marklund, Henrik and Haghgoo, Behzad and Ball, Robyn and Shpanskaya, Katie and Seekins, Jayne and Mong, David A. and Halabi, Safwan S. and Sandberg, Jesse K. and Jones, Ricky and Larson, David B. and Langlotz, Curtis P. and Patel, Bhavik N. and Lungren, Matthew P. and Ng, Andrew Y.},
title = {CheXpert: a large chest radiograph dataset with uncertainty labels and expert comparison},
year = {2019},
isbn = {978-1-57735-809-1},
publisher = {AAAI Press},
url = {https://doi.org/10.1609/aaai.v33i01.3301590},
doi = {10.1609/aaai.v33i01.3301590},
booktitle = {Proceedings of the Thirty-Third AAAI Conference on Artificial Intelligence and Thirty-First Innovative Applications of Artificial Intelligence Conference and Ninth AAAI Symposium on Educational Advances in Artificial Intelligence},
articleno = {73},
numpages = {8},
location = {Honolulu, Hawaii, USA},
series = {AAAI'19/IAAI'19/EAAI'19}
}

@article{gpt4o,
  title={Gpt-4o system card},
  author={Hurst, Aaron and Lerer, Adam and Goucher, Adam P and Perelman, Adam and Ramesh, Aditya and Clark, Aidan and Ostrow, AJ and Welihinda, Akila and Hayes, Alan and Radford, Alec and others},
  journal={arXiv preprint arXiv:2410.21276},
  year={2024}
}

@article{bootstrappingLLM, title={Bootstrapping Large Language Models for Radiology Report Generation}, volume={38}, url={https://ojs.aaai.org/index.php/AAAI/article/view/29826}, DOI={10.1609/aaai.v38i17.29826}, abstractNote={Radiology report generation (RRG) aims to automatically generate a free-text description from a specific clinical radiograph, e.g., chest X-Ray images. Existing approaches tend to perform RRG with specific models trained on the public yet limited data from scratch, where they often lead to inferior performance owing to the problem of inefficient capabilities in both aligning visual and textual features and generating informative reports accordingly. Currently, large language models (LLMs) offered a promising solution to text generation with their power in learning from big data, especially for cross-modal scenarios such as RRG. However, most existing LLMs are pre-trained on general data, and suffer from the same problem of conventional approaches caused by knowledge gap between general and medical domain if they are applied to RRG. Therefore in this paper, we propose an approach to bootstrapping LLMs for RRG with a in-domain instance induction and a coarse-to-fine decoding process. Specifically, the in-domain instance induction process learns to align the LLM to radiology reports from general texts through contrastive learning. The coarse-to-fine decoding performs a text elevating process for those reports from the ranker, further enhanced with visual features and refinement prompts. Experimental results on two prevailing RRG datasets, namely, IU X-Ray and MIMIC-CXR, demonstrate the superiority of our approach to previous state-of-the-art solutions. Further analyses illustrate that, for the LLM, the induction process enables it to better align with the medical domain and the coarse-to-fine generation allows it to conduct more precise text generation.}, number={17}, journal={Proceedings of the AAAI Conference on Artificial Intelligence}, author={Liu, Chang and Tian, Yuanhe and Chen, Weidong and Song, Yan and Zhang, Yongdong}, year={2024}, month={Mar.}, pages={18635-18643} }

@inproceedings{
vit,
title={An Image is Worth 16x16 Words: Transformers for Image Recognition at Scale},
author={Dosovitskiy, Alexey and Beyer, Lucas and Kolesnikov, Alexander and Weissenborn, Dirk and Zhai, Xiaohua and Unterthiner, Thomas and Dehghani, Mostafa and Minderer, Matthias and Heigold, Georg and Gelly, Sylvain and others},
booktitle={International Conference on Learning Representations},
year={2021},
url={https://openreview.net/forum?id=YicbFdNTTy}
}

@inproceedings{mca-rg,
  title={Mca-rg: Enhancing llms with medical concept alignment for radiology report generation},
  author={Xing, Qilong and Song, Zikai and Zhang, Youjia and Feng, Na and Yu, Junqing and Yang, Wei},
  booktitle={International Conference on Medical Image Computing and Computer-Assisted Intervention},
  pages={380--390},
  year={2025},
  organization={Springer}
}

\clearpage

\appendix

\section{Ablation Study}\label{subsec:ablation}

\begin{table*}[b]
\caption{
Results of the ablation study on the MIMIC-CXR all-image test set.
The table summarizes experiment outcomes based on text extraction level,
image encoder, extended settings, and RAG application.
${\cal L}_{\rm SC}$ represents semantic contrastive loss, and $\epsilon$ denotes
text embedding noise. Best values are highlighted in bold, and second-best
values are underlined.
}
\centering
\small
\setlength{\tabcolsep}{2.5pt}
\renewcommand{\arraystretch}{1.12}
\label{tab:Ablation}

\begin{adjustbox}{max width=\textwidth}
\begin{tabular}{c|c|c|c|c|ccc|c|ccc}
\Xhline{1pt}
\multicolumn{4}{c|}{Method}
& \multirow{2}{*}{Experiment}
& \multicolumn{3}{c|}{CheXbert}
& \multirow{2}{*}{\makecell{RadGraph \\ F1}}
& \multicolumn{3}{c}{NLG Metrics} \\
\cline{1-4} \cline{6-8} \cline{10-12}
Extraction Level & Image Encoder & Extended & RAG
& & micro-F1 & Macro-F1 & example-F1
& & ROUGE-L & BLEU-1 & BLEU-4 \\
\hline\hline

Sentence & \multirow{2}{*}{XrayDINOv2} & - & - & E1
& 57.3 & 37.7 & 48.9 & 24.3 & \textbf{26.1} & 37.9 & \textbf{10.1} \\
RadGraph Phrase & & - & - & E2
& 56.7 & 40.1 & 49.7 & 23.6 & 18.9 & 27.7 & 4.2 \\

\cline{1-12}
\multirow{10}{*}{Key Phrase}
 & XrayDINOv2 & - & - & E3
 & 57.2 & 41.2 & 49.5 & 25.6 & 22.3 & 36.0 & 7.3 \\
 & RAD-DINO & - & - & E4
 & 57.7 & 40.1 & 49.7 & 25.1 & 22.4 & 36.1 & 7.2 \\
 & XrayCLIP & - & - & E5
 & 57.4 & 41.0 & 49.5 & 25.7 & 23.0 & 36.2 & 7.3 \\
 & BiomedCLIP & - & - & E6
 & 47.0 & 26.0 & 38.6 & 20.8 & 20.9 & 30.6 & 4.9 \\
 & XrayDINOv2 + XrayCLIP & - & - & E7
 & 57.6 & 42.0 & 49.3 & 25.5 & 22.0 & 36.8 & 7.3 \\

\cline{2-12}
 & \multirow{3}{*}{XrayDINOv2 + XrayCLIP}
 & ${\cal L}_{\rm SC}$ & - & E8
 & 57.7 & 41.7 & 50.3 & 25.7 & 22.4 & 36.9 & 7.5 \\
 & & $\epsilon$ & - & E9
 & 58.3 & \textbf{42.5} & \underline{50.8} & 25.9 & 21.6 & 37.3 & 7.6 \\
 & & ${\cal L}_{\rm SC}$, $\epsilon$ & - & E10
 & \textbf{58.8} & \underline{42.3} & \textbf{51.1} & 25.7 & 23.5 & 36.2 & 7.4 \\

\cline{2-12}
 & \multirow{4}{*}{XrayDINOv2 + XrayCLIP}
 & \multirow{4}{*}{${\cal L}_{\rm SC}$, $\epsilon$}
 & Llama 3B & E11
 & 58.2 & 41.7 & 50.5 & 25.8 & 25.3 & \underline{38.3} & 8.0 \\
 & & & Llama 8B & E12
 & 58.5 & 42.0 & 50.7 & 26.3 & 24.7 & 36.7 & 7.2 \\
 & & & Llama 70B & E13
 & \underline{58.6} & 41.9 & 50.7 & \underline{26.6}
 & \underline{25.4} & \textbf{38.4} & \underline{8.2} \\
 & & & GPT-4o & E14 (RA-RRG)
 & 58.5 & 41.7 & 50.7 & \textbf{26.7} & 24.9 & 37.9 & 8.0 \\
\Xhline{1pt}
\end{tabular}
\end{adjustbox}
\end{table*}

To assess the effectiveness of our proposed method, we conducted ablation studies, as summarized in Table \ref{tab:Ablation}. 
For the experiments where RAG was not applied (from E1 to E10), the retrieved phrases were simply concatenated to form a single report for evaluation.
First, we examined the impact of text extraction levels used for training and retrieval.
E1, E2, and E3 were configured to segment a report by sentences, RadGraph extraction followed by rule-based graph construction, and the proposed key phrase extraction with an LLM, respectively.
E1, using full sentences, achieved the highest NLG metrics.
However, CheXbert Macro-F1 scores were the lowest for E1 and improved progressively across E2 and E3. 
E3 also achieved the best RadGraph F1 among the three experiments, 
suggesting that the proposed key phrase extraction was effective in enhancing clinical efficacy metrics.

Next, we examined the impact of different image encoders.
Experiments E3 to E6 employed single image encoders, while E7 applied multiple image encoders. 
Comparing DINOv2-based E3 and E4, other metrics showed minimal differences, but E3 (XrayDINOv2) demonstrated a noticeably higher Macro-F1.
Examining CLIP-based E5 and E6, E5 (XrayCLIP) showed significantly superior results. 
Consequently, in configuring the multiple image encoder for E7, we combined XrayDINOv2 and XrayCLIP using channel concatenation, yielding the highest Macro-F1 
of 42.0.

Finally, we assessed the impact of 
semantic contrastive loss and noise addition 
to text embedding, from E8 to E10.
Compared to E7, E8 improved example-F1 by 1.0 and RadGraph F1 by 0.2. 
E9 achieved the highest Macro-F1 (42.5) and increased RadGraph F1 by 0.4.
Applying both the loss and noise, E10 achieved the highest average CheXbert metrics,
indicating overall improvement.

Experiments from E11 to E14 used various sizes of Llama and GPT-4o as the LLM model in RAG, sharing the same retrieval model as E10.
Compared to E10, all experiments yielded higher RadGraph F1 scores, even for the smallest LLM (Llama 3B, E11).
NLG metrics improved as well, 
likely due to increased lexical similarity resulting from the 
LLM's ability to generate natural sentences from key phrases.
Comparing across E11 to E14,
neither the size nor the type of LLM had a significant impact on RRG performance,
suggesting that the critical factor is likely the key phrase retrieval rather than LLM performance.
Based on the ablation studies, we selected E14 with the best RadGraph F1 score of 26.7 as our final model.
Since we have empirically observed that randomness does not significantly impact the generation results, likely because all essential information is included in the prompt, we executed LLM inference only once.

\section{Key Phrase Extraction Details}\label{appendix:key_prhase}

\subsection{RadGraph Phrase Extraction}\label{radgraph-rule}

\begin{figure*}[ht!]
    \begin{center}
    \scalebox{0.95}{
        \includegraphics[width=\linewidth]{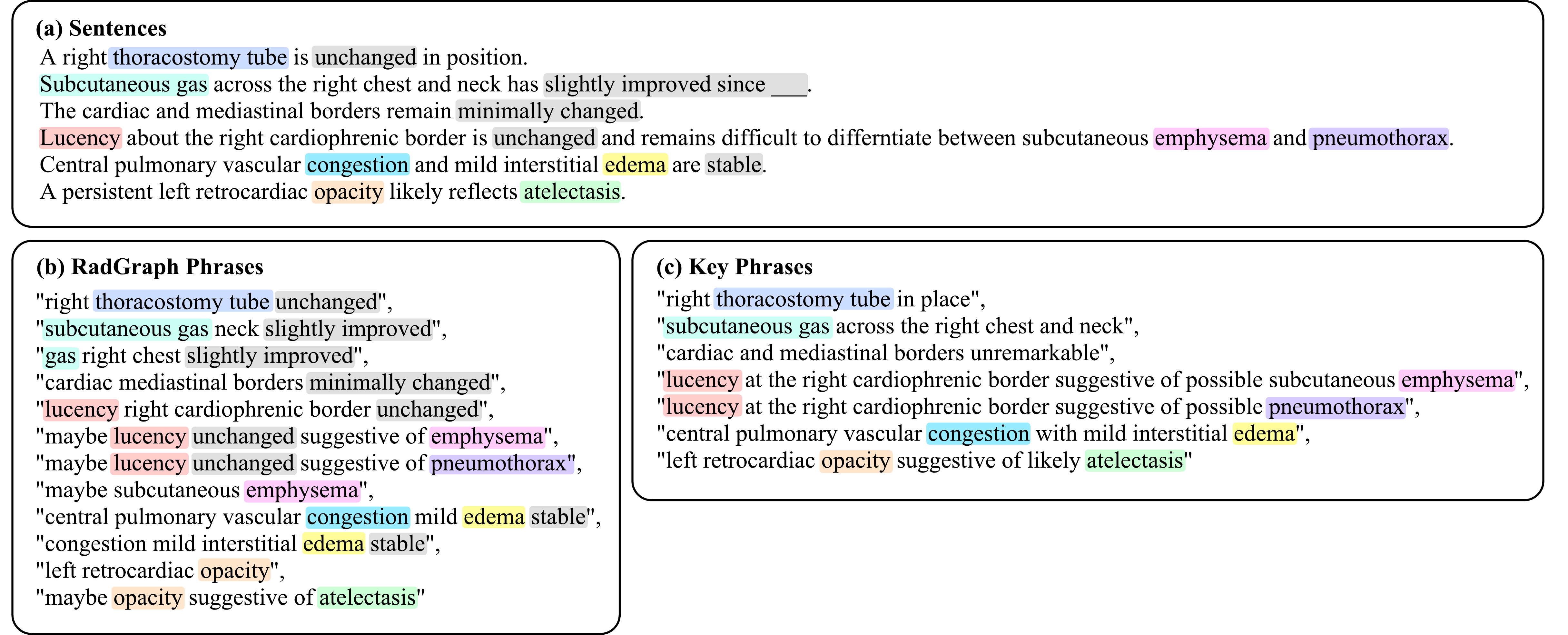}
    }
    \end{center}
    \vspace{-0.4cm}
    \caption{
    Example of retrieval target extraction from same radiology report as 
    (a) sentences, 
    (b) RadGraph phrases, and 
    (c) key phrases. 
    Key findings are highlighted using multiple colors, with the same color applied to identical findings. Phrases that may induce hallucinations are shown in gray.
    }
    \label{fig:comparison-data}

    \vspace{-0.4cm}

\end{figure*}

RadGraph extracts clinical entities and relations as a knowledge graph.
It captures three types of relations: `located\_at', `suggestive\_of', and `modify'.
To organize 
the extracted entities and these relations
into graphs representing minimal meaningful units, 
we apply the following rules: 
entities connected by `modify', which adds contextual meaning to another entity, are grouped within the same graph, 
while entities linked by `located\_at' and `suggestive\_of' are grouped into 
separate
graphs.
Each graph is then converted into a phrase.
For the three types of 
observation-related entities
(`OBS-DA': observation definitely absent, `OBS-DP': observation definitely present, and `OBS-U': observation uncertain),
we prepend `no' for `OBS-DA' and `maybe' for `OBS-U'.
Examples of the resulting phrases, 
referred to as `RadGraph phrases',
can be found in Figure \ref{fig:comparison-data} (b).

\subsection{LLM Prompt for Key Phrase Extraction}\label{subsec:llm-prompt}

\begin{figure}[ht!]
    \begin{center}
    \scalebox{0.95}{
        \includegraphics[width=\linewidth]{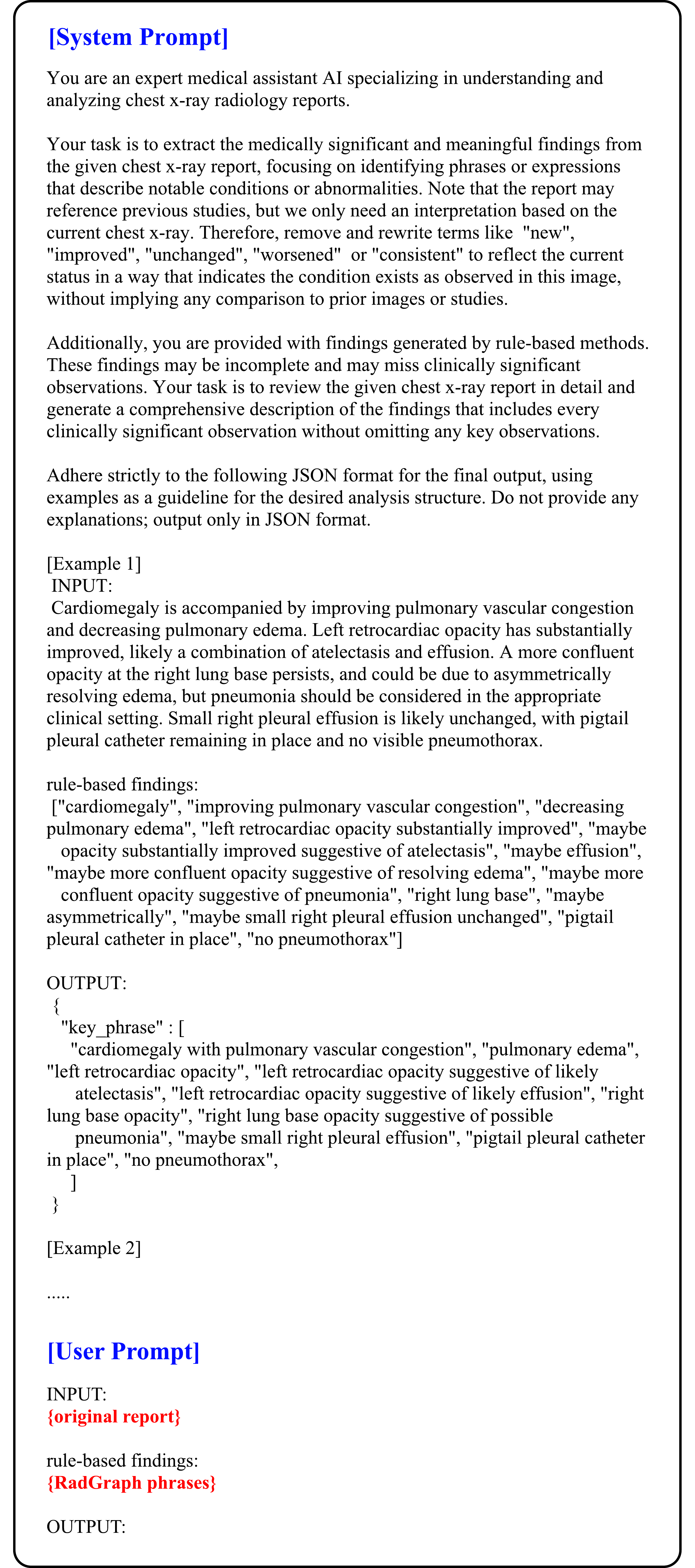}
    }
    \end{center}
    \vspace{-0.4cm}
    \caption{
    LLM prompt for key phrase extraction. The LLM extracts key phrases as a list by leveraging the original radiology report and RadGraph phrases.
    }
    \label{fig:key-phrase-extraction-prompt}
\end{figure}

The input prompt to the LLM
for key phrase extraction 
is designed 
to accurately extract clinically significant findings from radiology reports.
These findings are then organized into natural phrases that reflect
the current state.
As shown in Figure \ref{fig:key-phrase-extraction-prompt},
the input prompt instructs the LLM to identify key phrases based on the following guidelines.

First, the LLM is 
prompted to eliminate
comparative expressions such as 
``improved", ``unchanged", ``worsened" and ``consistent",
ensuring that
the extracted key phrases reflect 
only information directly 
inferable from the current image and
thereby minimizing hallucinations.
Since the LLM is a general-purpose model not specialized in the medical domain, 
it may miss clinically important details.
To mitigate this,
RadGraph phrases are included in the input prompt
alongside the original \textit{FINDINGS} section.
While these phrases may contain fragmented structures,
they help the LLM better capture clinically meaningful content.
Finally, 
representative examples of well-extracted key phrases are provided to more effectively
guide the model in extracting 
clinically relevant findings.

\subsection{Key Phrase Extraction Example}

Figure \ref{fig:comparison-data} shows three possible options for retrieval targets in retrieval-based RRG: 
(a) sentences from the \textit{FINDINGS} section,
(b) RadGraph phrases refined with rule-based processing after RadGraph extraction
and (c) the key phrases extracted from the proposed LLM prompting.
A comparison between Figure \ref{fig:comparison-data} (b) and Figure \ref{fig:comparison-data} (c) highlights the effectiveness of the LLM prompting described in Section \ref{subsec:llm-prompt}.

Figure \ref{fig:comparison-data} (b) includes past comparative expressions such as 
``unchanged" and ``improved"
(highlighted in gray) because these expressions appear in the original report, as shown in Figure \ref{fig:comparison-data} (a).
In contrast, Figure \ref{fig:comparison-data} (c) excludes such expressions, as the LLM was instructed to remove them.
Additionally, Figure \ref{fig:comparison-data} (b) contains multiple overlapping phrases representing a single finding,
such as ``emphysema" (highlighted in pink) and ``edema" (highlighted in yellow).
In Figure \ref{fig:comparison-data} (c), these overlapping phrases are combined
into a single key phrase that integrates all the scattered information, resulting in greater semantic clarity.
These observations demonstrate that LLM prompting is effective in minimizing potential hallucinations by removing past comparative expressions and in extracting clear and concise key phrases.

\section{LLM Prompt for RRG}

\begin{figure}[t!]
    \begin{center}
    \scalebox{0.88}{
        \includegraphics[width=\linewidth]{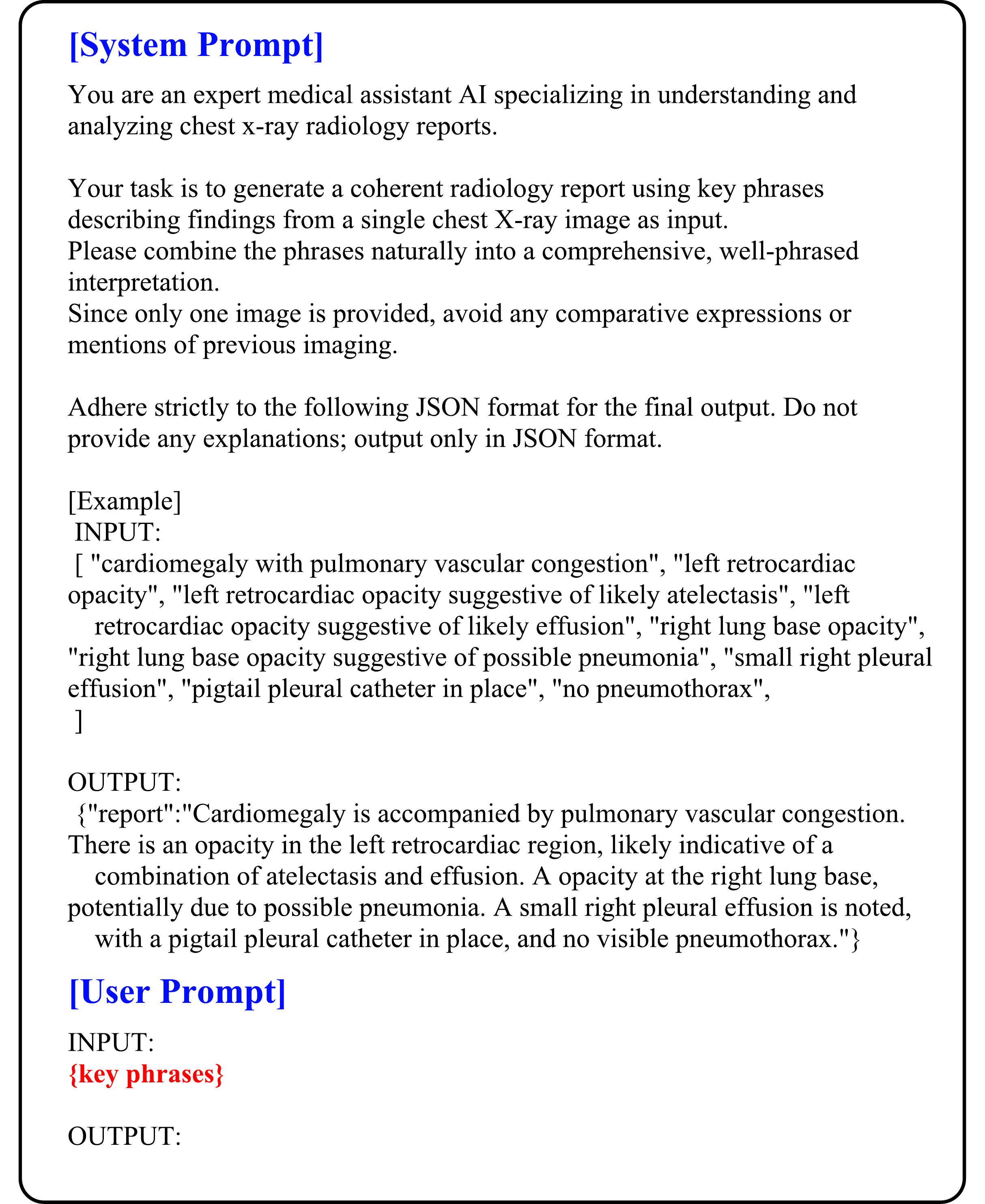}
    }
    \end{center}
    \vspace{-0.4cm}
    \caption{
    Single-view RAG prompt for RRG. Key phrases are provided as input to generate a radiology report.
    }
    \label{fig:RAG-prompt}
\end{figure}

\begin{figure}[b!]
    \begin{center}
    \scalebox{0.88}{
        \includegraphics[width=\linewidth]{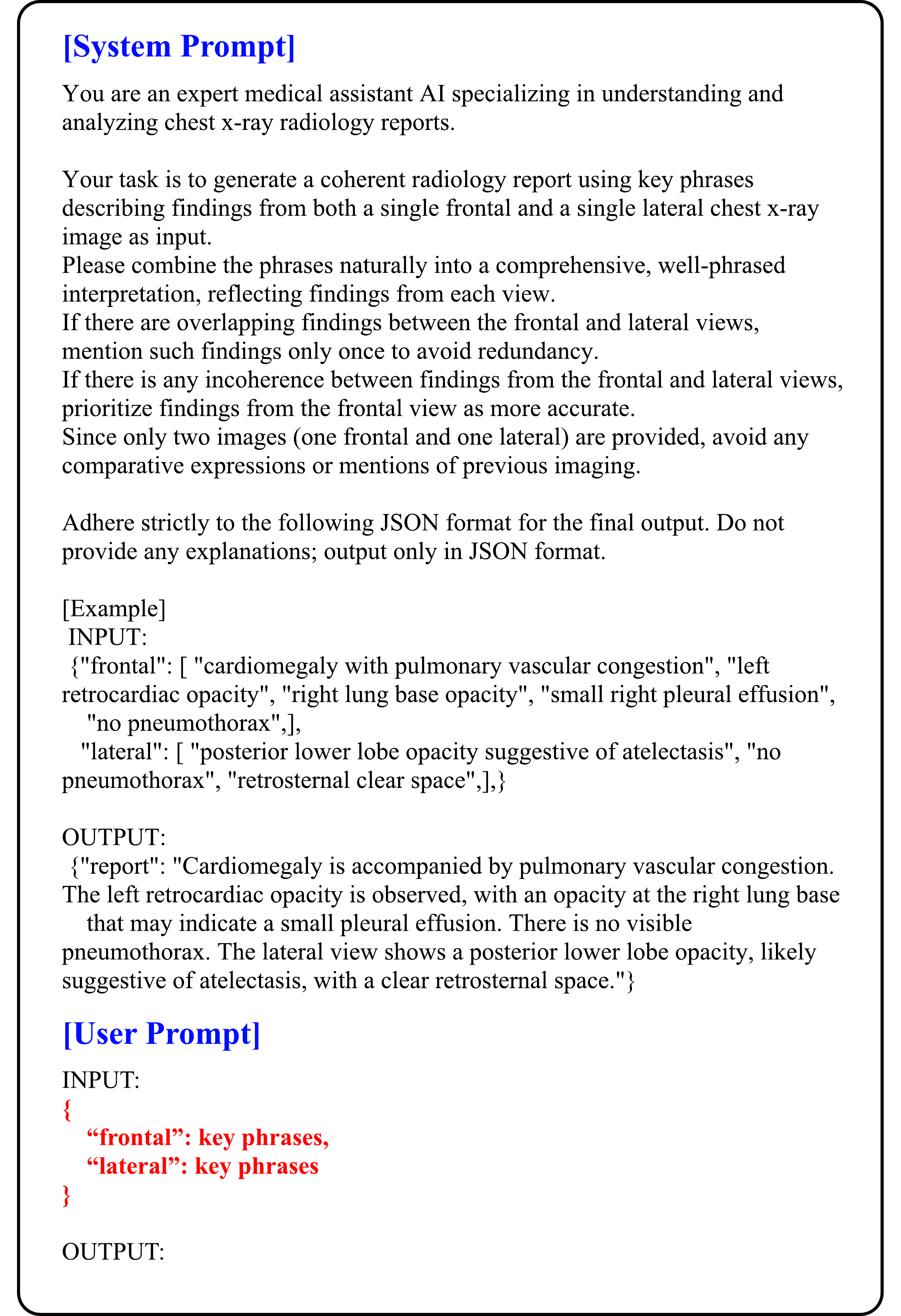}
    }
    \end{center}
    \vspace{-0.4cm}
    \caption{
Multi-view RAG prompt for RRG. Key phrases retrieved from the frontal and lateral images are separately provided as input to generate a radiology report.
    }
    \label{fig:multi-view-RAG-prompt}
\end{figure}

In the final step of generating the report with the LLM, an effective input prompt design is required to utilize the retrieved key phrases efficiently.
Figures \ref{fig:RAG-prompt} and \ref{fig:multi-view-RAG-prompt} illustrate the input prompts for the LLM in different contexts: 
Figure \ref{fig:RAG-prompt} shows the prompt used when a single
image, either frontal or lateral, is provided, 
whereas Figure \ref{fig:multi-view-RAG-prompt} illustrates the prompt for a two-view setting with both frontal and lateral images as input.
For both prompts, the LLM is required to remove any comparative expressions or references to prior study, as such expressions are definitively hallucinations given that only the current 
radiology data is
provided.

In Figure \ref{fig:multi-view-RAG-prompt}, additional instructions are provided to 
integrate the retrieved key phrases from the two different view images into a cohesive and natural report.
The system prompt directs the LLM to mention duplicate findings retrieved from both images only once.
For any conflicting phrases between the frontal and lateral view images, the retrieval result from the frontal view image takes priority.
This prioritization is based on the conventional perspective that the frontal view provides more critical information about the chest condition and includes more comprehensive diagnostic details compared to the lateral view.

\section{Additional Implementation Details}
\label{app:imp_details}
\paragraph{Vision Encoders}
To search for the best-performing vision encoder structure, we experiment with various 
CXR image encoders.
These include BiomedCLIP \cite{biomedclip} and XrayCLIP\footnote{\url{https://github.com/Stanford-AIMI/chexpert-plus}\label{fn:cxplus}} \cite{chexpertplus} for CLIP models, as well as RAD-DINO \cite{raddino} and XrayDINOv2\footref{fn:cxplus} \cite{chexpertplus} for DINOv2 models. 
Although XrayDINOv2 was 
originally 
trained at an image resolution of 224, we use a resolution of 518, interpolating positional embeddings as needed. 
For the final model, we combine multiple image encoders, specifically XrayDINOv2 and XrayCLIP. 
Since XrayDINOv2 has a longer visual token sequence, we interpolate XrayCLIP’s output and concatenate them channel-wise.

\paragraph{Text Encoder and Retriever}
For the text encoder we employ MPNet (`all-mpnet-base-v2')\footnote{\url{https://huggingface.co/sentence-transformers/all-mpnet-base-v2}} \cite{sbert} with 
embedding dimension $d=768$.
For the distribution-balanced loss ${\cal L}_{cls}$,
the hyperparameters are based on
COCO-MLT experimental settings.\footnote{\url{https://github.com/wutong16/DistributionBalancedLoss/blob/master/configs/coco/LT_resnet50_pfc_DB.py}}
The selection process is treated as single-label binary classification, with the positive class size set to 7.16 
(the average number of key phrases as described in Section
\ref{subsec:datasets}) 
and the negative class size fixed at 
$N - 7.16 = 42.84$. 

\paragraph{Optimization}
We use a learning rate of 0.0002 with a cosine decay scheduler and 50 warm-up steps. 
The retriever is trained
with a batch size of 128 for a maximum of 10 epochs, 
with the best model determined by
validation loss. 
Weight decay
is set to 0.05, and gradient clipping is applied with a 
maximum
value of 1.0. 
The optimizer is AdamW.
We train the 
model on a single H100 GPU for 18 hours, utilizing automatic mixed precision with bfloat16.

\paragraph{Large Language Models}
The LLMs used in this work are
`Llama-3.1-70B-Instruct'\footnote{\url{https://huggingface.co/meta-llama/Llama-3.1-70B-Instruct}} 
(abbreviated as Llama 70B 
\cite{llama3})
and GPT-4o\footnote{`gpt-4o-2024-08-06' through the OpenAI API}
\cite{gpt4o}.
For key phrase extraction (Section 
\ref{subsec:key_phrase_extraction}), 
radiology reports from the training data must be input into the LLM.
However, licensing restrictions for the training dataset (MIMIC-CXR)
explicitly prohibits sharing access to the data with third parties including sending it through APIs.
To address this,
we setup
the open-source Llama 70B model locally to generate LLM responses
instead.
The sampling parameters are set to their default values:
a temperature of 0.6 and a top $P$ probability of 0.9.
The vllm python package
\cite{vllm} 
is used with 4-bit quantization for inference.
In contrast, the final RRG step 
(Section \ref{subsec:mrg_llm})
inputs general medical key phrases (e.g., `no pleural effusion,' `mild cardiomegaly') into the LLM rather than full reports.
The key phrases extracted from reports are 
segmented and contain no
patient-specific information, 
allowing RAG experiments to be conducted 
using GPT-4o.
From a cost perspective,
approximately 485 reports are generated
for \$1, averaging \$0.002 per report.

\paragraph{Evaluation Tools}
We use publicly available implementations of standard evaluation metrics for the NLG metrics \footnote{\url{https://pypi.org/project/pycocoevalcap}} and RadGraph. \footnote{\url{https://github.com/rajpurkarlab/CXR-Report-Metric}}

\section{Label-Level Retrieval Error Analysis}\label{subsec:label-level-retrieval}

To further analyze retrieval errors at the label level, we evaluate the retrieval-only setting (without LLM generation) by computing CheXbert metrics for each of the 14 observation labels (Table~\ref{tab:label_level_retrieval}).
Common findings such as cardiomegaly and pleural effusion tend to have higher recall but lower precision, whereas rarer findings such as lung lesion and fracture exhibit high specificity but low recall, suggesting a tendency toward under-retrieval.

\begin{table}[ht!]
\caption{
Label-level CheXbert metrics in the retrieval-only setting (without LLM generation) on the MIMIC-CXR test set.
}
\centering
\small
\setlength{\tabcolsep}{2.5pt}
\renewcommand{\arraystretch}{1.12}
\label{tab:label_level_retrieval}

\begin{adjustbox}{max width=\columnwidth}
\begin{tabular}{l|c|c|c|c|c}
\Xhline{1pt}
Label & F1 & Recall & Precision & Specificity & Accuracy \\
\hline\hline
Enlarged Cardiomediastinum & 0.140 & 0.138 & 0.142 & 0.923 & 0.857 \\
Cardiomegaly & 0.703 & 0.844 & 0.602 & 0.684 & 0.742 \\
Lung Opacity & 0.612 & 0.681 & 0.556 & 0.675 & 0.678 \\
Lung Lesion & 0.237 & 0.191 & 0.313 & 0.969 & 0.915 \\
Edema & 0.512 & 0.744 & 0.390 & 0.782 & 0.776 \\
Consolidation & 0.176 & 0.176 & 0.176 & 0.957 & 0.918 \\
Pneumonia & 0.301 & 0.418 & 0.235 & 0.928 & 0.902 \\
Atelectasis & 0.551 & 0.713 & 0.448 & 0.688 & 0.695 \\
Pneumothorax & 0.382 & 0.615 & 0.277 & 0.967 & 0.960 \\
Pleural Effusion & 0.721 & 0.895 & 0.604 & 0.724 & 0.779 \\
Pleural Other & 0.235 & 0.204 & 0.277 & 0.978 & 0.948 \\
Fracture & 0.196 & 0.191 & 0.201 & 0.954 & 0.911 \\
Support Devices & 0.784 & 0.789 & 0.779 & 0.878 & 0.847 \\
No Finding & 0.371 & 0.397 & 0.349 & 0.953 & 0.920 \\
\Xhline{1pt}
\end{tabular}
\end{adjustbox}
\end{table}

\section{RAG Analysis}

\subsection{Retrieval Augmented RRG}

\begin{figure}[htbp]
    \begin{center}
    \scalebox{1.0}{
        \includegraphics[width=\linewidth]{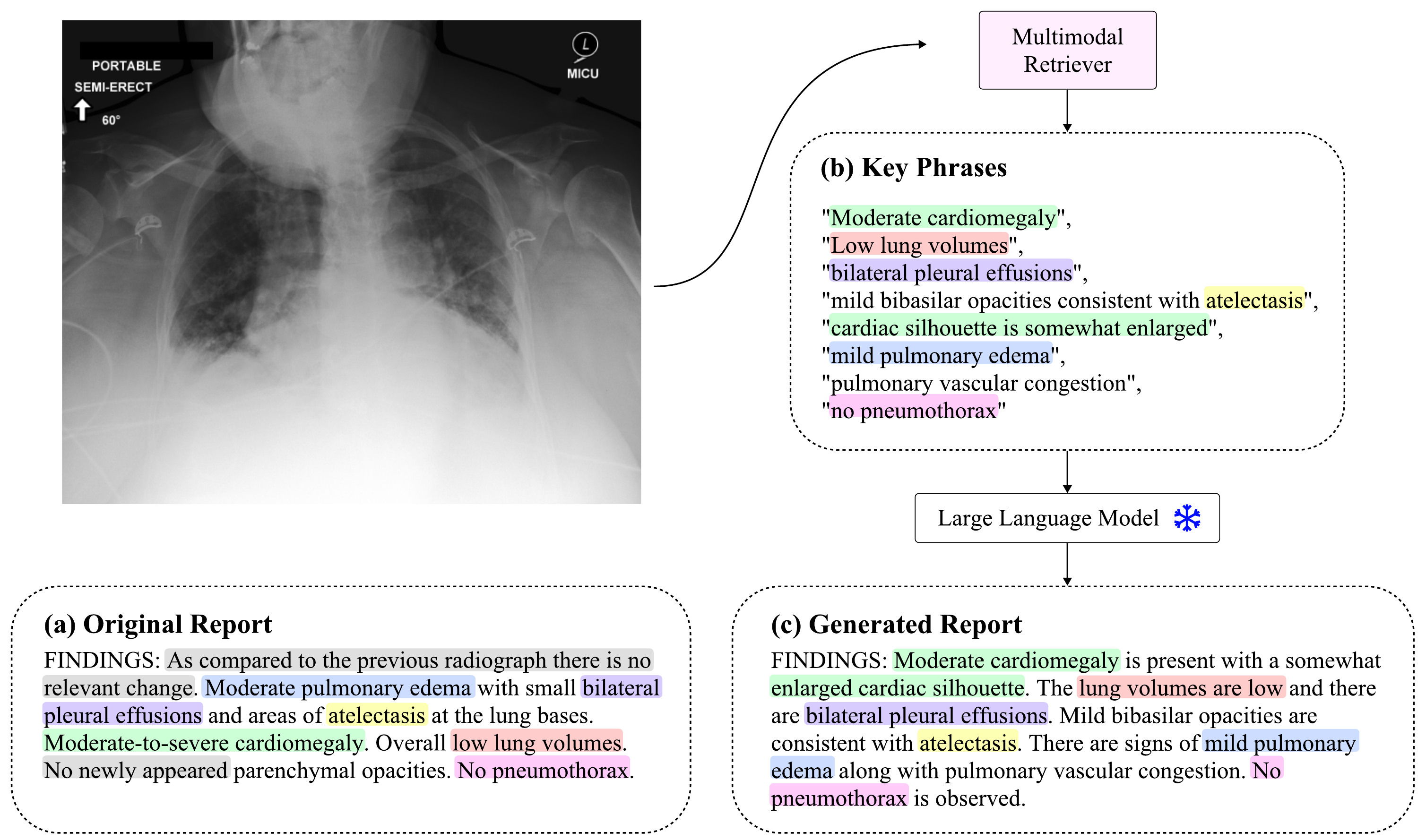}
    }
    \end{center}
    \vspace{-0.4cm}
    \caption{
    An example of key phrase retrieval results and the generated radiology report. Descriptions with the same meaning are highlighted in the same color, while content unsuitable for single-view RRG is shown in gray. 
    The sample is
    sourced from the MIMIC-CXR test set.
    }
    \label{fig:key-phrase-extraction}
\end{figure}

Figure \ref{fig:key-phrase-extraction} visualizes the process of RA-RRG leveraging 
a pre-trained LLM to generate a report
from the key phrases retrieved through multimodal retrieval.
Phrases that correspond to the same finding are highlighted in the same color.
Figure \ref{fig:key-phrase-extraction} (b) demonstrates that 
the key phrases derived from multimodal retrieval 
generally reflect the major findings in the original report
shown in Figure \ref{fig:key-phrase-extraction} (a).  
However, the phrase ``pulmonary vascular congestion," which is not explicitly mentioned in the original report, 
is added during the retrieval process, 
indicating false positive.
Figure \ref{fig:key-phrase-extraction} (c) illustrates how 
the LLM integrates the relationships between findings naturally 
and generates a structured and contextually coherent radiology report based on the input key phrases.
The generated report effectively incorporates the detailed information from the key phrases and preserves the major findings, 
consistent with
the original report.
Notably, the false positive phrase ``pulmonary vascular congestion'' was subsequently incorporated into the generated report.
This reveals a limitation of RA-RRG: it inherently propagates retrieval errors into the generated report, as its quality depends on the multimodal retriever.

\begin{table}[t]
\caption{Number of in-context examples for RRG.}
\centering
\small
\setlength{\tabcolsep}{3pt}
\renewcommand{\arraystretch}{1.1}
\label{tab:in_context}

\begin{adjustbox}{max width=\columnwidth}
\begin{tabular}{l|c|c|c|c|c}
\Xhline{1pt}
\multirow{2}{*}{In-context examples}
& \multirow{2}{*}{micro-F1}
& \multirow{2}{*}{Macro-F1}
& RadGraph
& \multirow{2}{*}{ROUGE-L}
& \multirow{2}{*}{BLEU-1} \\
& & & F1 & & \\
\hline\hline

0 example  & 58.4 & 41.6 & 26.5 & 24.5 & 35.2 \\
1 example  & 58.5 & 41.7 & 26.7 & 24.9 & 37.9 \\
3 examples & 58.2 & 41.6 & 26.7 & 25.2 & 38.2 \\
\Xhline{1pt}
\end{tabular}
\end{adjustbox}
\end{table}

\begin{figure}[t!]
    \begin{center}
    \scalebox{0.95}{
        \includegraphics[width=\linewidth]{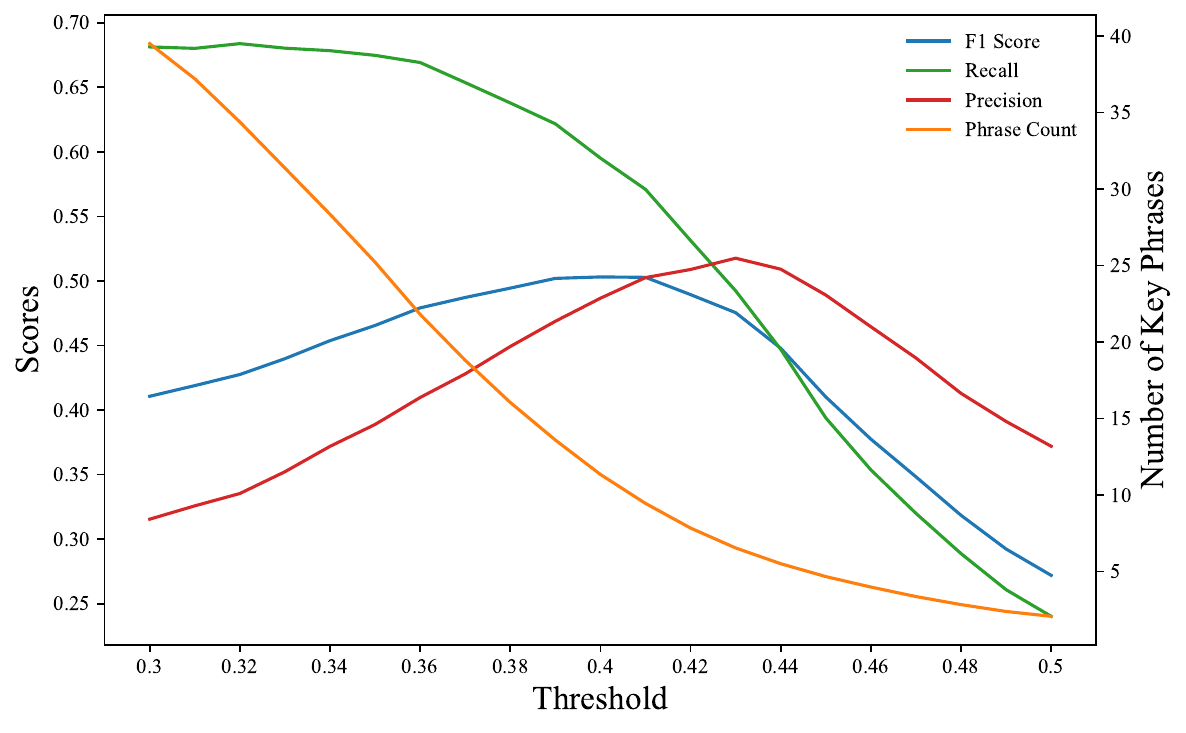}
    }
    \vspace{-0.5cm}
    \end{center}
    \caption{
    Impact of threshold on example-based average CheXbert scores and the number of key phrases.
    }
    \label{fig:threshold}
    \vspace{-0.5cm}
\end{figure}

\begin{figure*}[hb!]
    \begin{center}
    \scalebox{0.95}{
        \includegraphics[width=\linewidth]{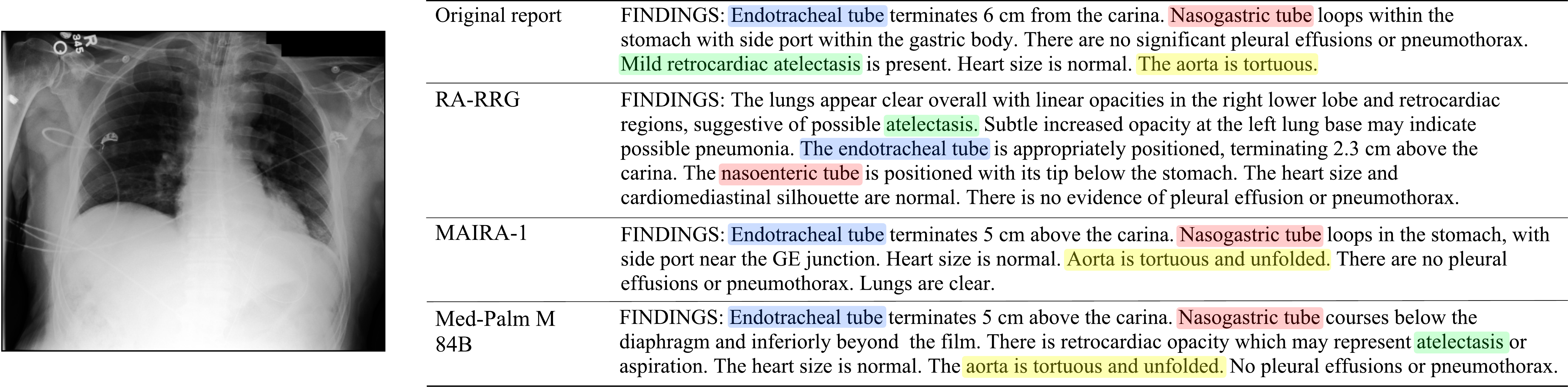}
    }
    \end{center}
    \vspace{-0.4cm} 
    \caption{
    Comparison of single-view RRG results. Positive findings are highlighted with different colors. 
    The sample is
    sourced from the MIMIC-CXR test set.
    Results for MAIRA-1 and Med-PaLM M 84B are 
    referenced from
    \citet{maira1} and \citet{medpalm-m}, respectively.
    }
    \label{fig:single-qualitative-examples_1}
\end{figure*}

\subsection{Effect of In-Context Example Quantity}

In Figure \ref{fig:RAG-prompt}, one in-context example is included in the prompt given as input to the LLM.
To assess the impact of the number of in-context examples, we vary the number of examples provided (0, 1, and 3).
Table \ref{tab:in_context} shows the results.
While more context examples for RRG slightly improved the NLP metrics, there was no gain in clinical efficacy. 
Considering the higher
cost of longer prompts,
we concluded that one example suffices for a clinically accurate report.

\subsection{Semantic Embedding Retrieval Threshold}

The number of retrieved key phrases in the inference stage is a crucial factor that directly influences the generated report.
This number varies for each image and is determined by the semantic embedding retrieval threshold. 
Figure \ref{fig:threshold} illustrates the average number of retrieved key phrases and the corresponding CheXbert example-based F1 score, precision, and recall across different thresholds.
The semantic embedding retrieval threshold of 0.4, which we set in Section \ref{subsec:impl_details}, is the value at which the example-based F1 score is maximized.

\section{Qualitative Examples}

Figure \ref{fig:single-qualitative-examples_1} 
presents a comparison of
the radiology reports generated by RA-RRG, MAIRA-1, and Med-PaLM M 84B based on the findings in the original report.
RA-RRG generally 
captured the findings mentioned in the original report well, 
particularly by providing clear descriptions of the positions of the ``endotracheal tube'' and ``nasogastric tube'' and addressing ``atelectasis'' appropriately.  
However, it 
omitted phrases
such as ``the aorta is tortuous'' and 
introduced details absent from
the original report, such as ``subtle increased opacity at the left lung base may indicate possible pneumonia''.
This demonstrates RA-RRG's ability to reflect key findings while occasionally including unnecessary details.  
MAIRA-1 also performed well in addressing the findings from the original report but missed ``atelectasis'' and inaccurately described the side port location of the ``nasogastric tube,'' showing limitations in certain details.  
Med-PaLM M 84B generally 
addressed most
findings accurately but incorrectly described the position of the ``NG tube'' as extending beyond the film.

\begin{figure*}[ht!]
    \begin{center}
    \scalebox{0.95}{
        \includegraphics[width=\linewidth]{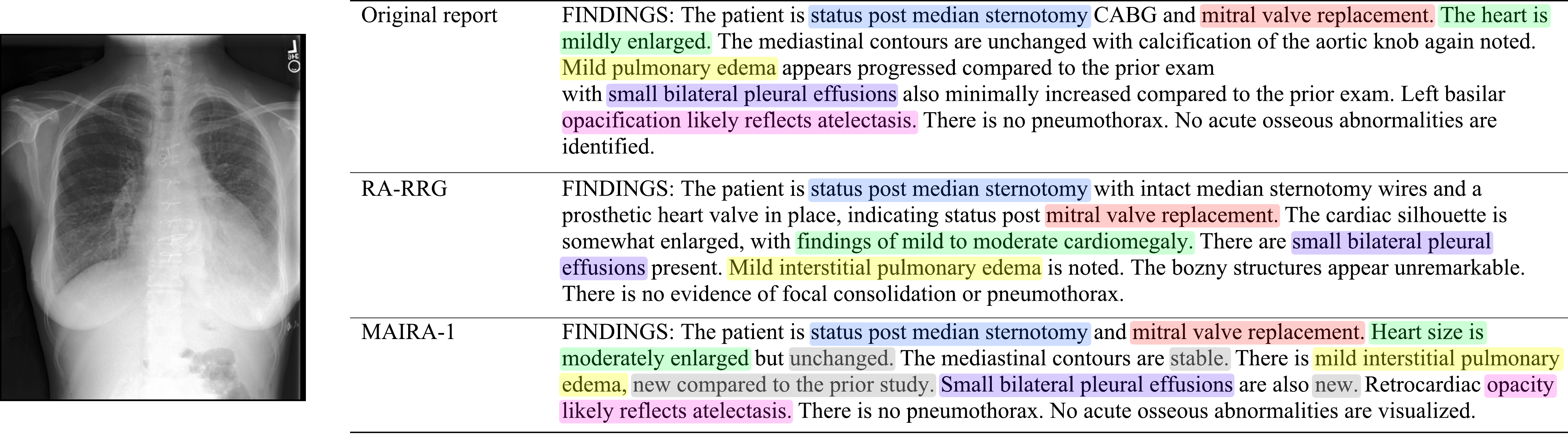}
        }
    
    \end{center}
    \vspace{-0.4cm}
    \caption{
    Comparison of single-view RRG results. Positive findings are highlighted with different colors, and phrases considered to be hallucinations are shown in gray.
    The sample is
    sourced from the MIMIC-CXR test set.
    MAIRA-1's result is from \citet{maira1}.
    }
    
    \label{fig:single-qualitative-examples_2}
\end{figure*}

\begin{figure*}[ht!]
    \begin{center}
    \scalebox{0.95}{
        \includegraphics[width=\linewidth]{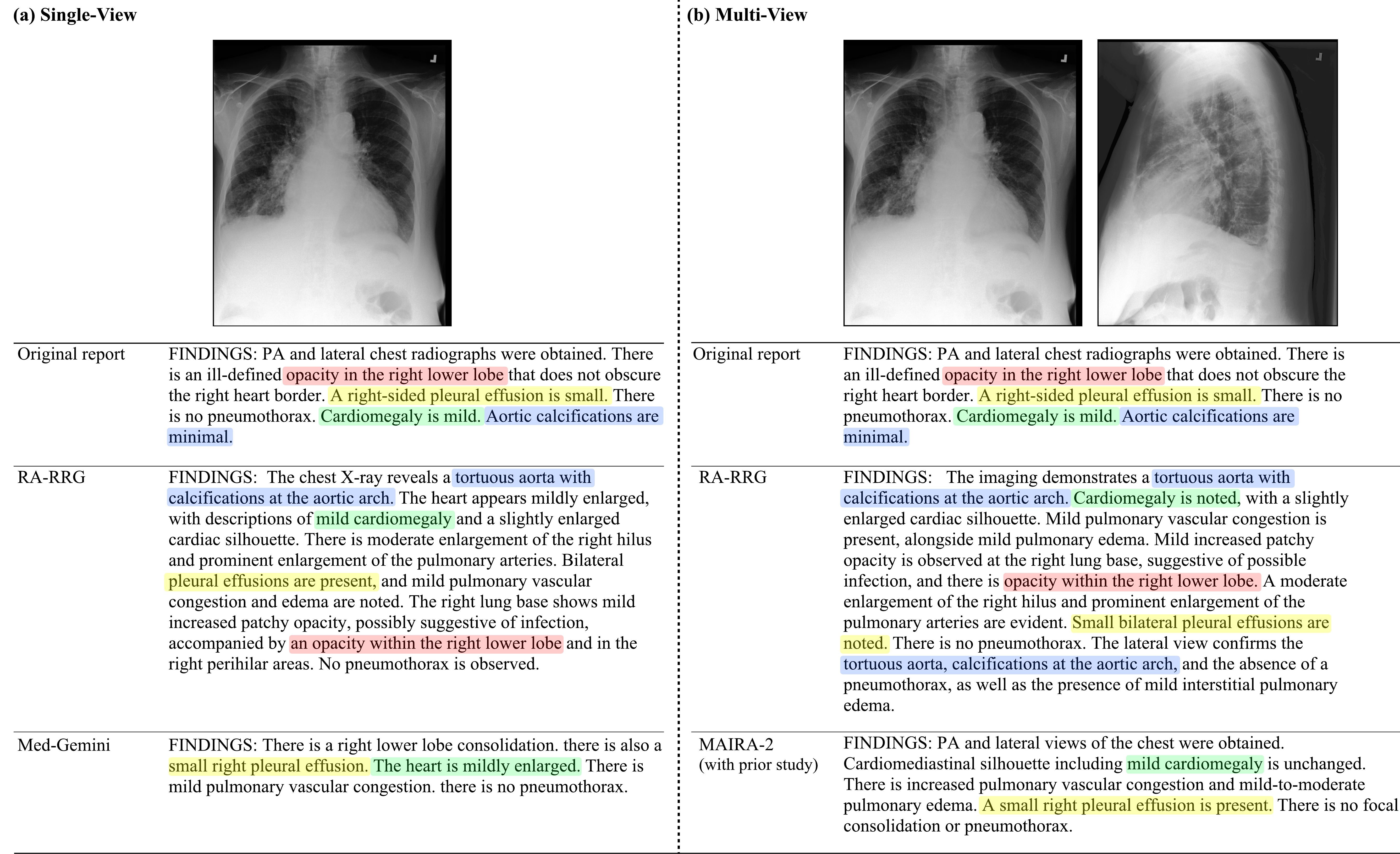}
    }
    \end{center}
    \vspace{-0.4cm}
    \caption{
Comparison of (a) single-view and (b) multi-view RRG results for the same study.
The report for MAIRA-2 was generated using multi-view inputs along with additional prior study information.
Positive findings are highlighted with different colors. 
The sample is
sourced from the MIMIC-CXR validation set. 
Results for Med-Gemini \cite{medgemini2d} and MAIRA-2 \cite{maira2} are referenced from their respective papers.
    }
    \label{fig:multi-qualitative-examples}
\end{figure*}

Figure \ref{fig:single-qualitative-examples_2} illustrate how accurately RA-RRG and MAIRA-1 identify the key findings from the given CXR image.
RA-RRG 
missed
findings such as ``opacification likely reflects atelectasis'' and ``calcification''.
However, it generally captured 
other key findings appropriately.
In contrast, MAIRA-1 effectively captured the key findings but shared the same limitation in failing to mention ``calcification.''  
Additionally, it exhibited hallucinations, such as including comparisons to prior studies that do not align with the single-view RRG or referencing unnecessary changes.

Figure \ref{fig:multi-qualitative-examples} compares the results of RA-RRG, Med-Gemini, and MAIRA-2 for the same study, with each model performing RRG under different input scenarios.
Figure \ref{fig:multi-qualitative-examples} (a) compares the outcomes  
of RA-RRG and Med-Gemini on a single frontal view image.
RA-RRG generally 
reflected the original report's key findings,
but it also 
added observations not present in the source,
such as ``moderate enlargement of the right hilus'' and ``prominent enlargement of the pulmonary arteries.''  
It also showed inconsistency with the original report by describing the location of ``pleural effusion'' as ``bilateral,'' whereas the original report indicated ``right-sided.''  
In contrast, Med-Gemini failed to mention key findings such as ``opacity in the right lower lobe'' and ``aortic calcifications,'' which 
are interpreted as
significant omissions of critical pathological information. 
Additionally, Med-Gemini 
introduced unnecessary details
not included in the original report, such as ``mild pulmonary vascular congestion.''

Figure \ref{fig:multi-qualitative-examples} (b) 
displays the comparison of 
RA-RRG and MAIRA-2 after adding the lateral view from the same study as the frontal view in Figure \ref{fig:multi-qualitative-examples} (a).  
It is worth noting that
the radiology report of MAIRA-2 was generated using multi-view inputs along with additional prior study data.
As a result, the generated results of MAIRA-2 in Figure \ref{fig:multi-qualitative-examples} (b) include comparative expressions referencing the past, but these are not considered hallucinations and are therefore not highlighted in gray in the figure.
RA-RRG, similar to its result in Figure \ref{fig:multi-qualitative-examples} (a), exhibited errors in the location of ``pleural effusion'' and generated additional 
details absent from
the original report.  
Meanwhile, MAIRA-2 failed to mention ``right lower lobe opacity'' and ``aortic calcification'' and was observed 
adding extra content
not included in the original report, such as ``pulmonary vascular congestion'' and ``mild-to-moderate pulmonary edema.''

RA-RRG demonstrated competitive performance with state-of-the-art MLLMs without requiring LLM fine-tuning and showed seamless adaptability to multi-view RRG. 
Additionally, the use of key phrase extraction and RAG appears to effectively suppress hallucinations. 
However, compared to the original reports, some false positives with additional descriptions and false negatives from missed findings were observed, highlighting the need for further 
improvements.

\end{document}